\documentclass[10pt,twocolumn,letterpaper]{article}

\usepackage{cvpr}
\usepackage{times}
\usepackage{epsfig}
\usepackage{graphicx}
\usepackage{amsmath}
\usepackage{amssymb}
\usepackage{xcolor}
\usepackage{colortbl}
\usepackage{multirow}
\usepackage{multicol}
\usepackage{array}
\usepackage{verbatim}
\usepackage{booktabs}
\usepackage{cite}
\usepackage{enumitem}

\usepackage[super]{nth}
\usepackage[subrefformat=parens]{subcaption}
\usepackage[group-separator={,},range-phrase=--]{siunitx}

\usepackage[pagebackref=true,breaklinks=true,letterpaper=true,colorlinks,bookmarks=false]{hyperref}

\usepackage[capitalise]{cleveref}

\graphicspath{{Figures/}{Figures/Graphs/}}

\begin{document}

%%%%%%%%% TITLE
\title{\vspace{-3mm} % TODO
Privacy-enhancing Sclera Segmentation Benchmarking Competition: SSBC 2025\vspace{-2.5mm}}

\author{%
\small%
M.~Vitek$^{1,}$\thanks{M.~Vitek and D.~Tomašević are first authors with equal contributions.}\;, D.~Tomašević$^{1,*}$, A.~Das$^2$,
S.~Nathan$^4$,
G.~\"Ozbulak$^{5,6}$, G.~A.~T.~\"Ozbulak$^{7,8}$, J.~P.~Calbimonte$^{8,9}$, A.~Anjos$^{5,6}$,\\[-1mm]\small%
H.~H.~Bhatt$^{10}$, D.~D.~Premani$^{10}$, J.~Chaudhari$^{10}$,
C.~Wang$^{11}$, J.~Jiang$^{12}$, C.~Zhang$^{11}$, Q.~Zhang$^{12}$,
I.~I.~Ganapathi$^{13}$, S.~S.~Ali$^{13}$,\\[-1mm]\small{}D.~Velayudan$^{13}$, M.~Assefa$^{13}$, N.~Werghi$^{13}$,
Z.~A.~Daniels$^{14}$,
L.~John$^{15}$, R.~Vyas$^{15}$,
J.~N.~Khiarak$^{16}$, T.~A.~Saeed$^{17}$, M.~Nasehi$^{18}$,\\[-1mm]\small{}A.~Kianfar$^{19}$, M.~P.~Panahi$^{20}$,
G.~Sharma$^{21}$, P.~R.~Panth$^{21}$, R.~Ramachandra$^{22}$, A.~Nigam$^{21}$,
U.~Pal$^3$, P.~Peer$^1$, V.~Štruc$^1$\\
\scriptsize%
$^1$University of Ljubljana (UL, SI), $^2$Birla Institute of Technology \& Science Pilani (BITS Pilani, IN), $^3$Indian Statistical Institute (ISI, IN),
$^4$Couger Inc.\ (JP),\\[-1.5mm]\scriptsize%
$^5$\'Ecole Polytechnique F\'ed\'erale de Lausanne (EPFL, CH), $^6$Idiap Research Institute (CH), $^7$Universit\'e de Lausanne (UNIL, CH),\\[-1.5mm]\scriptsize{}$^8$University of Applied Sciences and Arts Western Switzerland (HES-SO, CH), $^9$The Sense Innovation and Research Center (CH),
$^{10}$Ahmedabad University (AU, IN)\\[-1.5mm]\scriptsize%
$^{11}$Beijing University of Civil Engineering and Architecture (BUCEA, CN), $^{12}$People’s Public Security University of China (PPSUC, CN),\\[-1.5mm]\scriptsize%
$^{13}$Khalifa University of Science and Technology (KU, AE),
$^{14}$SRI International (US),
$^{15}$Pandit Deendayal Energy University (PDPU, IN),\\[-1.5mm]\scriptsize%
$^{16}$Warsaw University of Technology (WUT, PL), $^{17}$Pirogov Russian National Research Medical University (RNRMU, RU),\\[-1.5mm]\scriptsize{}$^{18}$Amirkabir University of Technology (AUT, IR), $^{19}$Institute for Advanced Studies in Basic Sciences (IASBS, IR), $^{20}$Seraj Institute (SI, IR),\\[-1.5mm]\scriptsize%
$^{21}$Indian Institute of Technology Mandi (IIT-M, IN), $^{22}$Norwegian University of Science and Technology (NTNU, NO)\vspace{-3mm}%
}

\maketitle
\thispagestyle{empty}

%%%%%%%%%%%%%%%%%%%%%%%%%%%%%%%%%%%%%%%%%%%%%%%%%%%%%%%%%%%%%%%%%%%%%%%%%%%%%%%%%%%%%
\begin{abstract}\vspace{-2mm}
%%%%%%%%%%%%%%%%%%%%%%%%%%%%%%%%%%%%%%%%%%%%%%%%%%%%%%%%%%%%%%%%%%%%%%%%%%%%%%%%%%%%%

This paper presents a summary of the 2025 Sclera Segmentation Benchmarking Competition (SSBC), which focused on the development of privacy-preserving sclera-segmentation models trained using synthetically generated ocular images.~The goal of the competition was to evaluate how well models trained on synthetic data perform in comparison to those trained on real-world datasets. The competition featured two tracks: $(i)$ one relying solely on synthetic data for model development, and $(ii)$ one combining/mixing synthetic with (a limited amount of) real-world data. A total of nine research groups submitted diverse segmentation models, employing a variety of architectural designs, including transformer-based solutions, lightweight models, and segmentation networks guided by generative frameworks. Experiments were conducted across three evaluation datasets containing both synthetic and real-world images, collected under diverse conditions. Results show that models trained entirely on synthetic data can achieve competitive performance, particularly when dedicated training strategies are employed, as evidenced by the top performing models that achieved $F_1$ scores of over $0.8$ in the synthetic data track. Moreover, performance gains in the mixed track were often driven more by methodological choices rather than by the inclusion of real data, highlighting the promise of synthetic data for privacy-aware biometric development. 
The code and data for the competition is available at: \url{https://github.com/dariant/SSBC_2025}.
\vspace{-4mm}%The findings highlight the potential of synthetic data to support privacy-conscious research in biometric segmentation without significant compromises in accuracy.
%The paper presents a summary of the 2025 Sclera Segmentation Benchmarking Competition (SSBC), the 9th in the series of group benchmarking efforts centered around the problem of sclera segmentation. Different from previous editions, the goal of SSBC 2020 was to evaluate the performance of sclera-segmentation models on images captured with mobile devices. The competition was used as a platform to assess the sensitivity of existing models to i) differences in mobile devices used for image capture and ii) changes in the ambient acquisition conditions. 26 research groups registered for SSBC 2020, out of which 13 took part in the final round and submitted a total of 16 segmentation models for scoring. These included a wide variety of deep-learning solutions as well as one approach based on standard image processing techniques. Experiments were conducted with three recent datasets. Most of the segmentation models achieved relatively consistent performance across images captured with different mobile devices (with slight differences across devices), but struggled most with low-quality images captured in challenging ambient conditions, i.e., in an indoor environment and with poor lighting.
\end{abstract}

%%%%%%%%%%%%%%%%%%%%%%%%%%%%%%%%%%%%%%%%%%%%%%%%%%%%%%%%%%%%%%%%%%%%%%%%%%%%%%%%%%%%%
\section{Introduction}
%%%%%%%%%%%%%%%%%%%%%%%%%%%%%%%%%%%%%%%%%%%%%%%%%%%%%%%%%%%%%%%%%%%%%%%%%%%%%%%%%%%%%

% \begin{figure}[t]
% \centering
%   \includegraphics[width=\textwidth]{Figures/Teaser.pdf}
% \caption{\color{blue}The quality of ocular images captured by mobile devices depends heavily on the imaging sensor used and the ambient acquisition conditions present during the capturing process - as illustrated by the sample images above. The goal of SSBC 2020 is to evaluate the performance of different sclera segmentation models across different capturing devices and acquisition conditions in the largest group benchmarking effort in this problem domain so far.}
% \label{fig: teaser}
% \end{figure}

\begin{figure}[t!]
    \centering
    \includegraphics[width=0.83\linewidth]{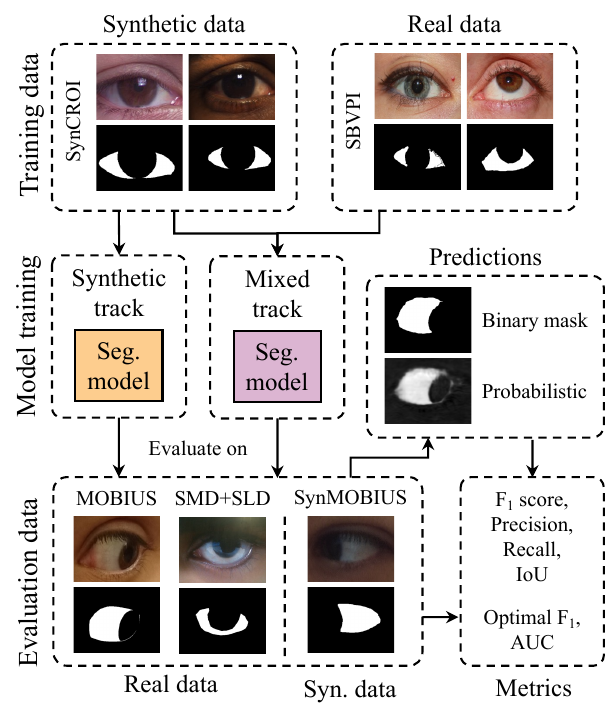}\vspace{-2mm}
    \caption{
    \textbf{Overview of SSBC 2025}.~The competition entailed two tracks, where sclera segmentation models were trained on \textit{(i)} synthetic data and \textit{(ii)} a mix of synthetic and real data. Models were evaluated on sequestered real and synthetic datasets. Participants were required to submit binary and probabilistic predictions.\vspace{-4mm}}
    \label{fig:teaser}
\end{figure}

Ocular biometrics represent a popular branch of research that focuses on computer-aided techniques, capable of inferring the identity of individuals based on distinctive ocular traits.   
%Ocular biometrics is a field within biometric recognition that focuses on identifying individuals through distinctive eye-related features. 
While iris recognition has historically dominated this field, recent research is increasingly looking into other ocular characteristics that can either complement or substitute the iris.~Among these, the sclera, the white region of the eye, has emerged as a promising candidate due to its distinctiveness, long-term consistency, and resistance to presentation attacks \cite{das2013scleras,rot2020deep,vitek2020comprehensive}. Unlike the iris, the sclera is less susceptible to degradations from common visual conditions and can be imaged under normal lighting with standard cameras, making it especially practical for real-world applications.

Research into sclera biometrics has intensified in recent years, covering a wide range of topics including recognition algorithms \cite{zhou2011new, riccio2017unsupervised, das2016framework, das2013sclera, alkassar2015robust, vitek2020comprehensive, das2021efficient, das2022sclera, vitek2025gazenet}, segmentation methods \cite{rot2020deep, radu2015robust, alkassar2016sclera, vitek2022exploring, vitek2023ipad}, detection of presentation attacks \cite{rot2020deep, vitek2020comprehensive, das2016framework}, adaptability to user variability \cite{das2015online, das2014new}, fusion techniques \cite{das2017decision, gottemukkula2016method}, and lightweight methods \cite{garbin2019openeds, vitek2023ipad, vitek2025gazenet}. However, with research in this area generally moving towards data-hungry deep learning models, %which require large amounts of data to train on, 
a key issue that has arisen is privacy preservation, since large-scale datasets of ocular images compiled with privacy-protecting measures in mind are largely unavailable.~A possible solution to this issue is the use of synthetically generated data \cite{das2017sclera,tomavsevic2022bioculargan,tomavsevic2024bifacegan,tomavsevic2024generating,tomavsevic2025idbooth}, which maintains the characteristics required to develop biometric models, but does not belong to (or contain identifying information about) real-world individuals and is, thus, not open to abuse or breach of privacy.

To investigate the effectiveness of synthetic data for sclera biometrics, the 2025 Sclera Segmentation Benchmarking Competition (SSBC), summarized in \cref{fig:teaser}, was organized in the scope of the International Joint Conference on Biometrics (IJCB 2025).~The competition focused on sclera segmentation, which is a crucial step in any sclera recognition pipeline, as the segmentation quality directly impacts all subsequent stages, including normalization, feature extraction, and identity matching.~The competition aimed at answering key research questions, such as: How well does synthetic data mimic real-world datasets? What effect does the use of synthetic training data have on segmentation performance? What techniques can be used to adapt models to the use of synthetically generated images during training? Nine segmentation models submitted by visible research teams from around the world were evaluated to provide insights into these and related questions. The joint effort of the organizers and the participating teams resulted in the following contributions:
\begin{itemize}[topsep=0pt,parsep=0pt,itemsep=0pt]
    \item A comprehensive evaluation of contemporary sclera segmentation models on real-world and synthetically generated evaluation data.
    \item A  differential performance analysis, studying the impact of the use of synthetic and real-world training data in model development.    
    \item A study of the impact of model size and computational complexity on the final segmentation performance and on the model's adaptability to synthetic training data.
\end{itemize}

%%%%%%%%%%%%%%%%%%%%%%%%%%%%%%%%%%%%%%%%%%%%%%%%%%%%%%%%%%%%%%%%%%%%%%%%%%%%%%%%%%%%%
\section{Related Work}

SSBC 2025 is the \nth{9} iteration of the sclera segmentation benchmarking competition, originally started at the BTAS conference in 2015. The SSBC series of competitions has significantly pushed forward the development of sclera segmentation models, with each iteration addressing a different research problem.~The \nth{1} and \nth{3} SSBC (SSBC 2015 and SSBC 2016), studied the segmentation performance of various models and additionally introduced new datasets for sclera segmentation (i.e., MASD and SMD) \cite{sserbc15}.~The \nth{2} iteration (SSRBC 2016) studied recognition approaches in addition to sclera segmentation techniques \cite{das2016ssrbc}.~The \nth{4} iteration, SSERBC 2017, again included the recognition task, but additionally explored the impact of gaze direction on successful segmentation and recognition~\cite{sserbc17}. The \nth{5} competition, SSBC 2018, studied the impact of cross-sensor image capture on the performance of sclera segmentation~\cite{sserbc18}, while the \nth{6} iteration, SSBC 2019, investigated how cross-resolution environments affect segmentation performance~\cite{sserbc19}. The \nth{7} edition, SSBC 2020, introduced a novel dataset (MOBIUS), compiled specifically for mobile sclera biometrics, and, consequently, focused on sclera segmentation in the mobile domain~\cite{ssbc2020}. A follow-up effort to SSBC 2020 \cite{vitek2022exploring} then explored various types of biases present in contemporary sclera segmentation models.~Finally, the \nth{8} iteration (SSRBC 2023) looked into segmentation and recognition performance individually, as well as the interplay between the two tasks  \cite{ssbc2023}.

Differently from past SSBC editions, SSBC 2025 focuses on privacy-preserving sclera segmentation models, developed with the use of synthetically generated (identity-less) training data. The main (distinct) goal of the competition is to study how well such models perform in relation to models trained on real-world data from actual individuals.
\begin{table}[t]
\renewcommand{\arraystretch}{1.1}
\caption{Summary of the real-world and synthetic datasets used for SSBC 2025. Reported is the number of images and subjects, the main sources of variability and the purpose in the competition.}
\label{tab:databases}
\centering
\resizebox{\columnwidth}{!}{%
\begin{tabular}{lrrrrr}
\toprule
Origin & Dataset & \#Images & \#IDs    &  Variability & Purpose\\
\toprule
\multirow{3}{*}{{Real-world}} & SBVPI & $1840$ & $55$ &  GZ, CLR & Training\\
& SMD+SLD & $489$ & $52$ & CN & Testing \\
& MOBIUS & $3540$ & $35$ & GZ, CLR, CN & Testing\\
\midrule
\multirow{3}{*}{{Synthetic}} & SynCROI (CE) & $5500$ & N/A & GZ, CLR & Training \\
 & SynCROI (PU) & $5500$ & N/A & CN & Training \\
 & SynMOBIUS & $4772$ & N/A  &  GZ, CLR, CN & Testing \\
\bottomrule
\multicolumn{6}{l}{(GZ) - gaze, (CLR) - eye color, (CN) - acquisition condition}\\% \TODO{Blur?}}\\
\vspace{-8mm}
\end{tabular}}
\end{table}

\begin{figure*}[t]
    \centering
    % --- Top row: SBVPI and SMD+SLD ---
    \begin{subfigure}{.44\linewidth}
        \includegraphics[width=2.5cm,height=1.5cm]{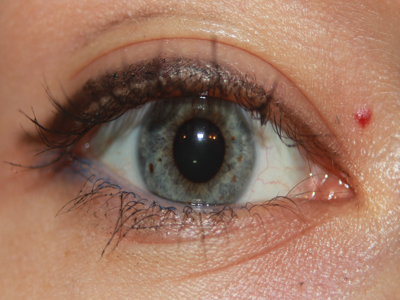}\hfill%
        \includegraphics[width=2.5cm,height=1.5cm]{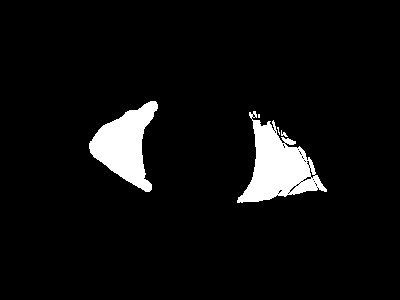}\hfill%
        \includegraphics[width=2.5cm,height=1.5cm]{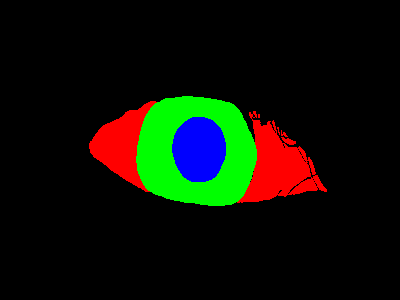}\\%\vspace{1mm}\\
        \includegraphics[width=2.5cm,height=1.5cm]{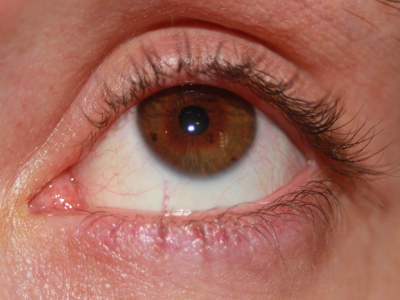}\hfill%
        \includegraphics[width=2.5cm,height=1.5cm]{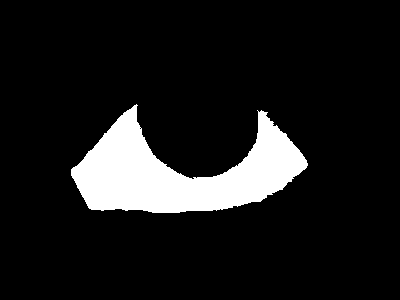}\hfill%
        \includegraphics[width=2.5cm,height=1.5cm]{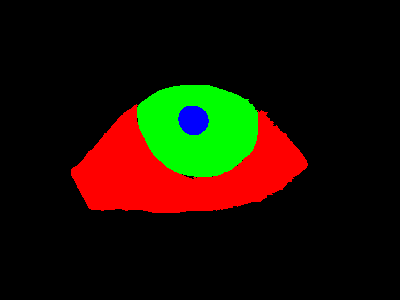}\hfill
        \caption{SBVPI}
        \label{fig:sbvpi}
    \end{subfigure}%\hfill%
    \hspace{4mm}
    \begin{subfigure}{.44\linewidth}
        \includegraphics[width=2.5cm,height=1.5cm]{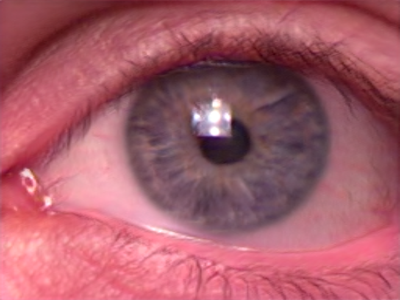}\hfill%
        \includegraphics[width=2.5cm,height=1.5cm]{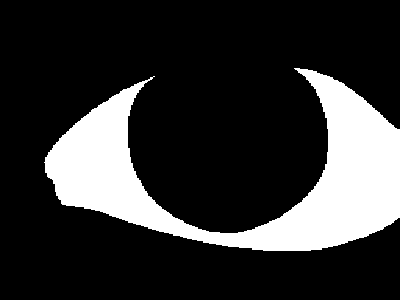}\hfill%
        \includegraphics[width=2.5cm,height=1.5cm]{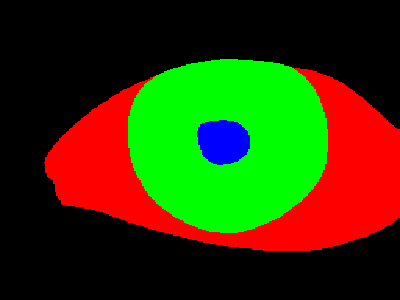}\\%\vspace{1mm}\\
        \includegraphics[width=2.5cm,height=1.5cm]{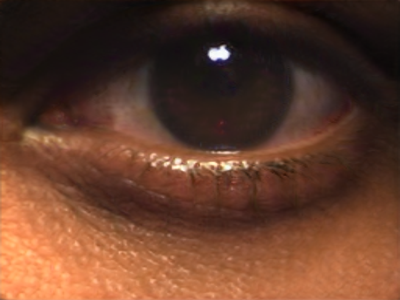}\hfill%
        \includegraphics[width=2.5cm,height=1.5cm]{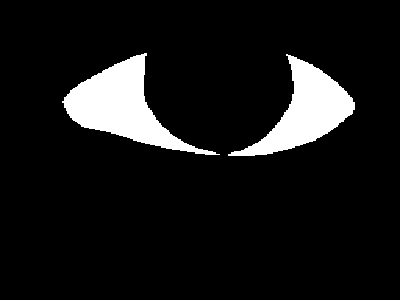}\hfill%
        \includegraphics[width=2.5cm,height=1.5cm]{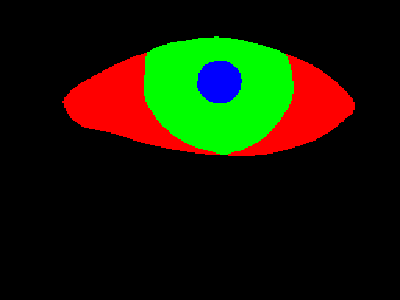}
        \caption{SynCROI}
        \label{fig:syncroi}
    \end{subfigure}
    \vskip 1mm  % Adjust this for vertical spacing between top and bottom groups
    % --- Bottom row: MOBIUS, SMD+SLD, SynMOBIUS ---
    \begin{subfigure}{.29\linewidth}
        \includegraphics[width=2.5cm,height=1.5cm]{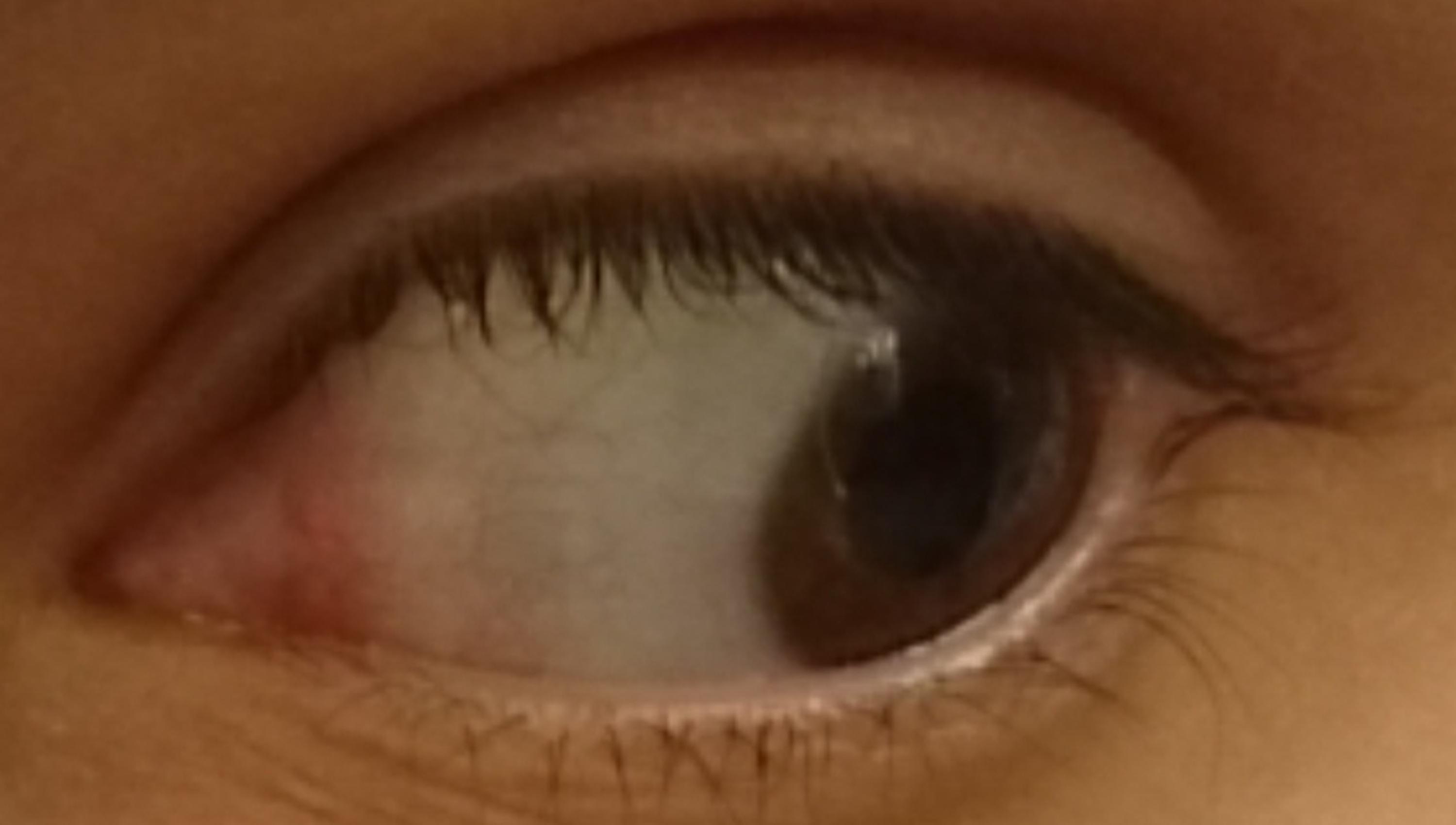}\hfill% 
        \includegraphics[width=2.5cm,height=1.5cm]{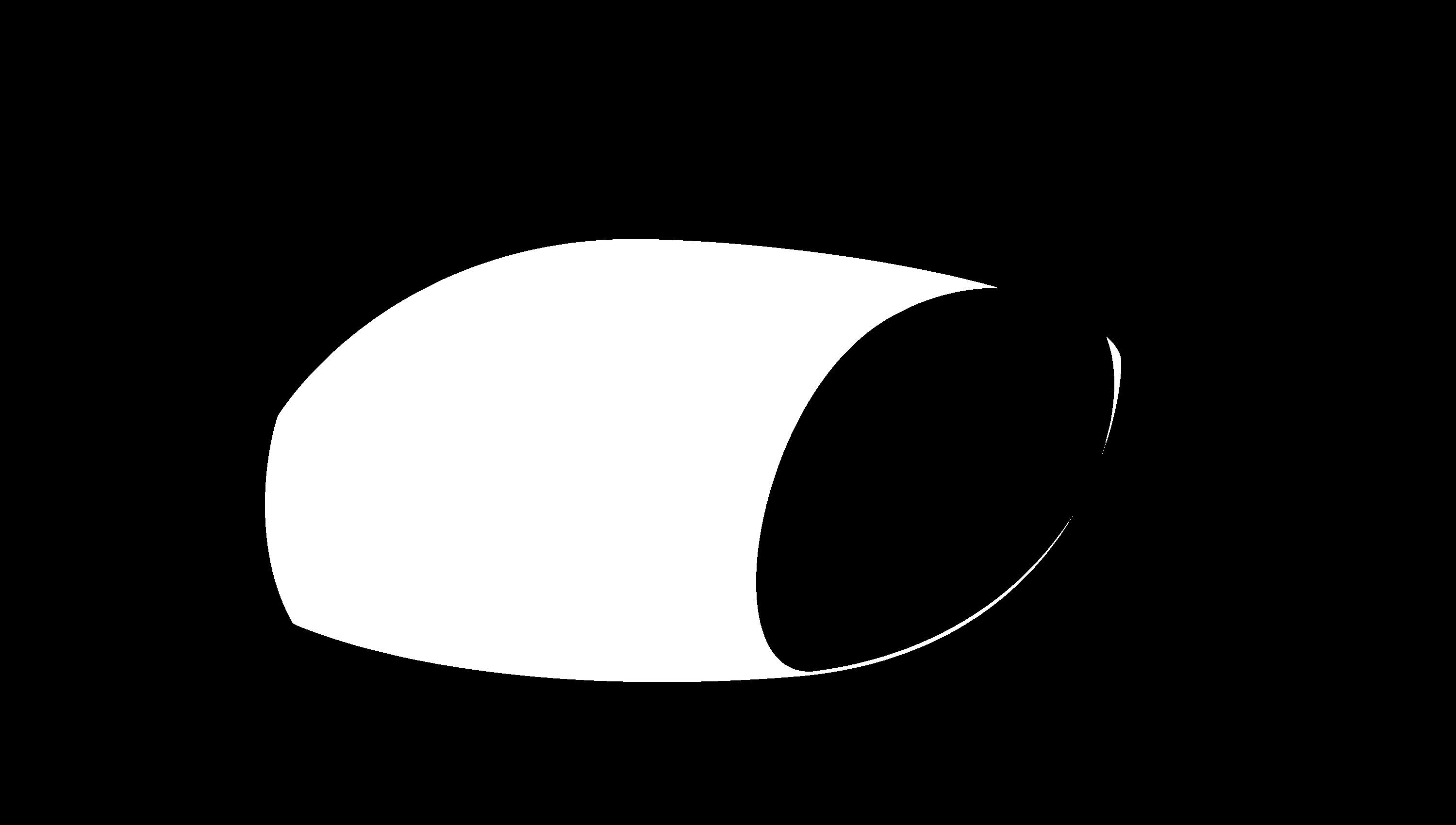}\\
        \includegraphics[width=2.5cm,height=1.5cm]{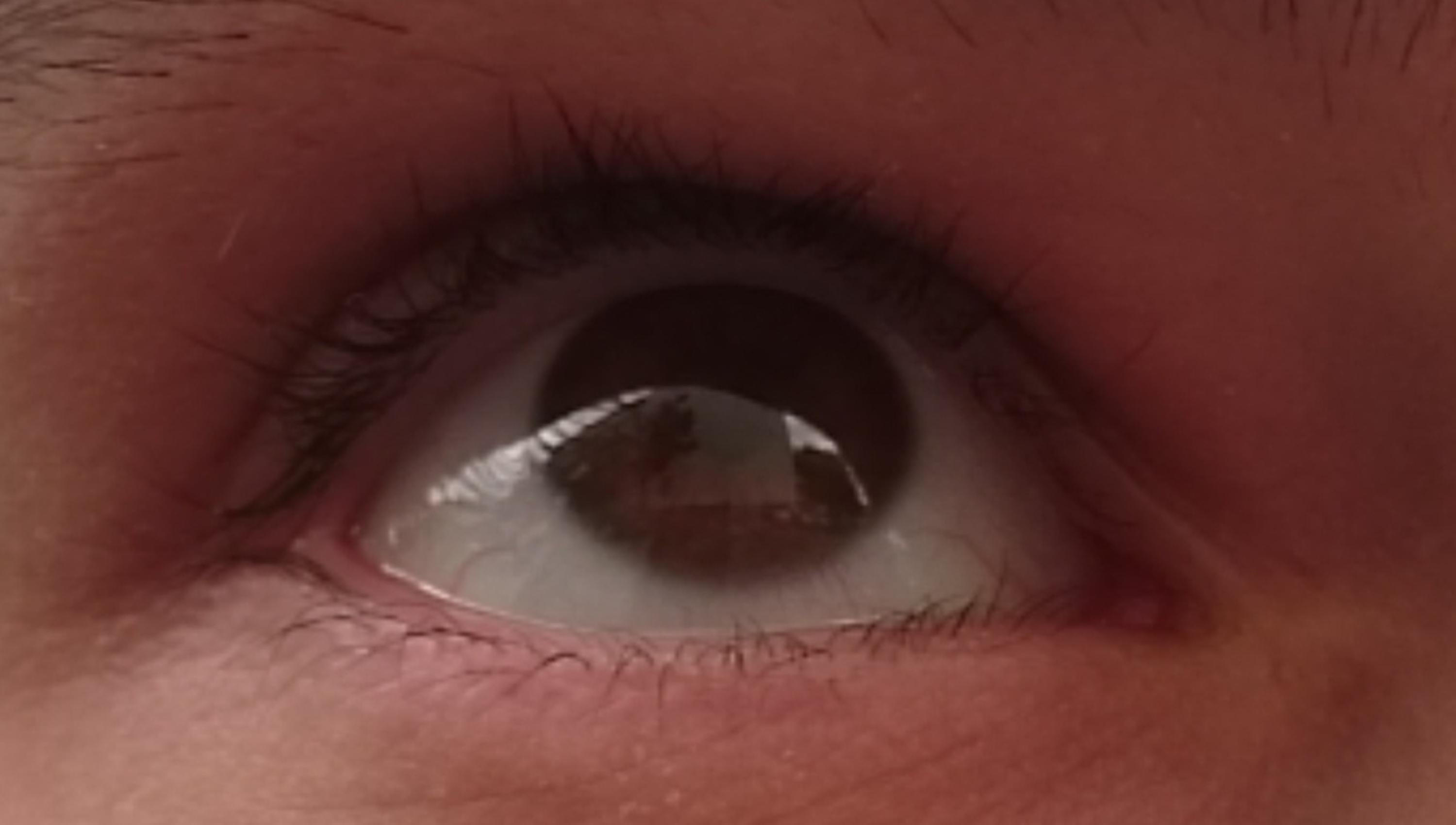}\hfill%
        \includegraphics[width=2.5cm,height=1.5cm]{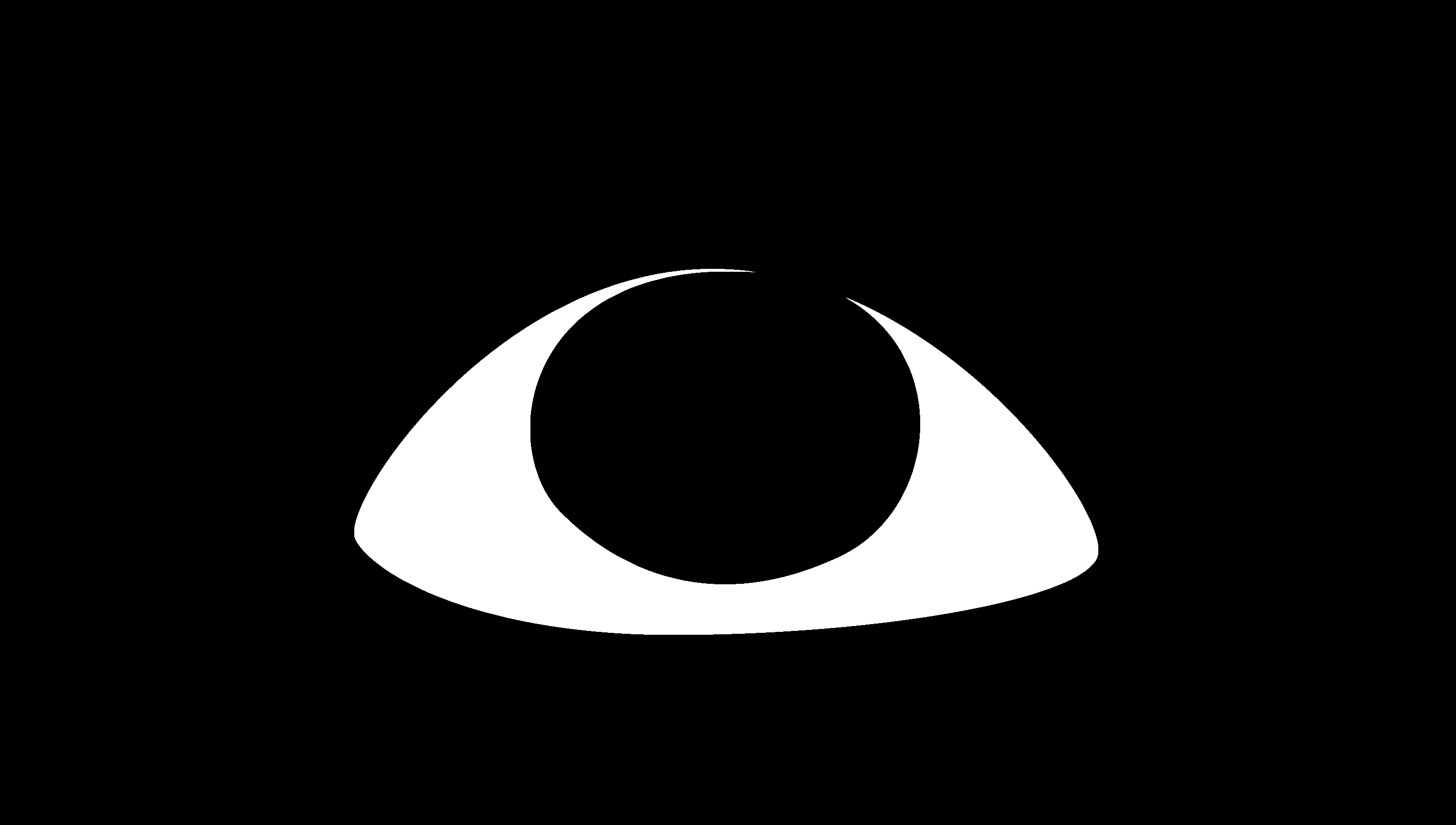}
        \caption{MOBIUS}
        \label{fig:mobius}
    \end{subfigure}
    \hspace{4mm}
    \begin{subfigure}{.29\linewidth}
        \includegraphics[width=2.5cm,height=1.5cm]{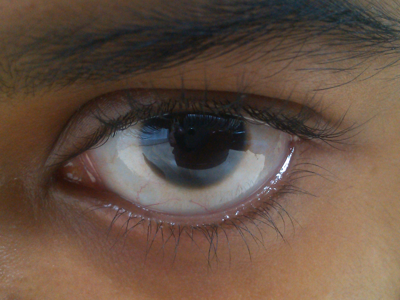}\hfill%
        \includegraphics[width=2.5cm,height=1.5cm]{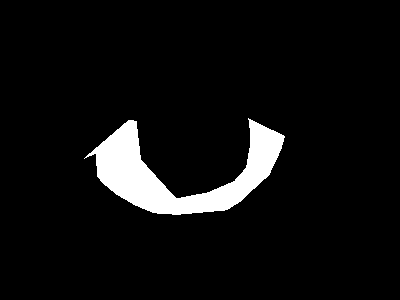}\\
        \includegraphics[width=2.5cm,height=1.5cm]{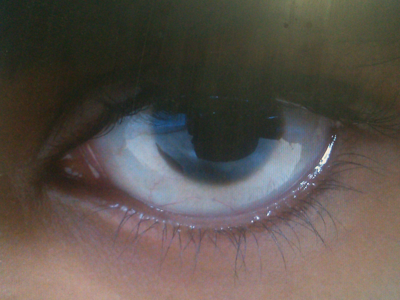}\hfill%
        \includegraphics[width=2.5cm,height=1.5cm]{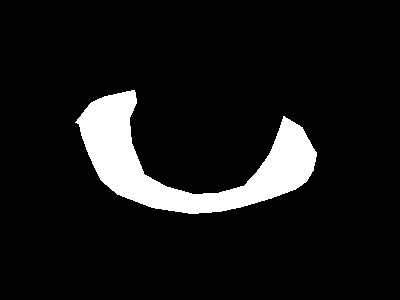}
        \caption{SMD+SLD}
        \label{fig:smdsld_test}
    \end{subfigure}
    \hspace{4mm}
    \begin{subfigure}{.29\linewidth}
        \includegraphics[width=2.5cm,height=1.5cm]{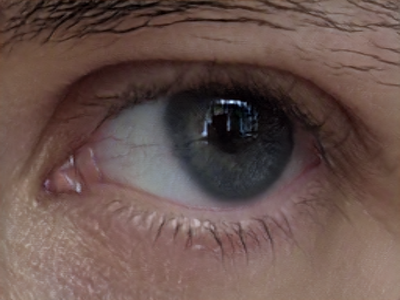}\hfill% 
        \includegraphics[width=2.5cm,height=1.5cm]{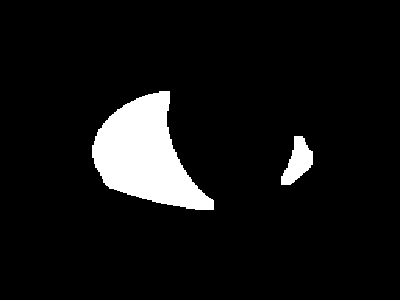}\\
        \includegraphics[width=2.5cm,height=1.5cm]{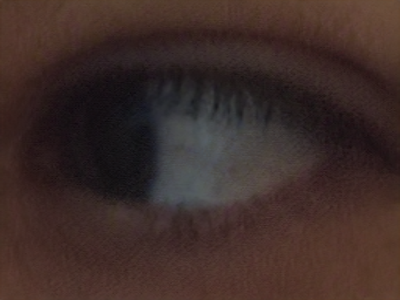}\hfill%
        \includegraphics[width=2.5cm,height=1.5cm]{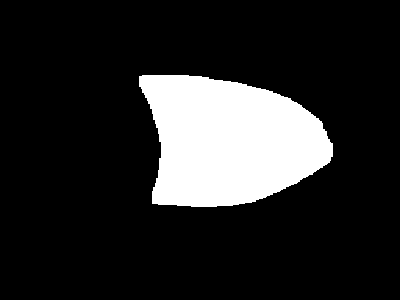}
        \caption{SynMOBIUS}
        \label{fig:synmobius}
    \end{subfigure}
    %\vspace{-1mm}
    \caption{%{\color{blue} Lahko tudi obdržimo samo binary maske (ceprav so imeli na voljo tudi vse regije) ali pa dodamo več slik / jih drugače razporedimo / jim spremenimo ratio.} 
    \textbf{Samples from the training and testing datasets used in SSBC 2025}.~The top row contains ocular images and available ground truth segmentation masks of the training datasets, while the bottom row depicts testings samples and their reference masks. %Note that the training data also contains a 4-class mark-up. 
    SSBC 2025 participant were free to train their models either on the complete 4-class mark-up, or only on the region corresponding to the sclera.
    } \vspace{-4mm}
    \label{fig:datasets}
\end{figure*}

%%%%%%%%%%%%%%%%%%%%%%%%%%%%%%%%%%%%%%%%%%%%%%%%%%%%%%%%%%%%%%%%%%%%%%%%%%%%%%%%%%%%%
\section{SSBC 2025 Competition Data}
\label{sec:methodology}
%{\color{blue}Bi imeli ločen chapter za generiranje sintetične podatke? Ali samo section}

The aim of SSBC is to study the use of synthetic data for training and testing of sclera segmentation models and investigate what impact synthetic data has on model performance when compared to real-world sclera imagery. To this end, several real-world (R) and synthetic datasets (S) were used for the competition, i.e., 
%Differently from previous iterations of the competition, SSBC 2025 focuses on the integration of both real-world and synthetic datasets, including: 
\textit{(i)} the {Sclera Blood Vessels, Periocular and Iris} (SBVPI) dataset (R)~\cite{vitek2020comprehensive}, \textit{(ii)} the {Mobile Ocular Biometrics in Unconstrained Settings} (MOBIUS) dataset (R)~\cite{ssbc2020},  \textit{(iii)} the combined Sclera Mobile Dataset~\cite{das2017towards} and Sclera Liveness Dataset (R)~\cite{ssbc2023} (i.e., SMD+SLD), as well as \textit{(iv)} the {Synthetic Cross-Racial Ocular Image} (SynCROI) dataset (S), and \textit{(v)} the {Synthetic MOBIUS} (SynMOBIUS) dataset (S). 
SBVPI and SynCROI represent the primary training datasets used throughout SSBC 2025, while MOBIUS, SMD+SLD, and SynMOBIUS were used for evaluation purposes only. 
A summary of the datasets is provided in Table \ref{tab:databases}, while sample images of each dataset and their corresponding segmentation masks are presented in Figure \ref{fig:datasets}. As part of SSBC 2025, images of all datasets were resized to a resolution of $400 \times 300$. 

% Detailed descriptions of the real-world and synthetic datasets can be found in the following sections, along with the competition protocol and performance metrics.

% This sections 

% In this section, we present the benchmarking methodology of SSBC 2025. We begin with the presentation of real and synthetic datasets utilized for training and evaluating segmentation models as part of the challenge. We then delve into the evaluation protocol of the competition and the used performance metrics.    

%%%%%%%%%%%%%%%%%%%%%%%%%%%%%%%%
\subsection{Real--World Datasets}\label{sec:datasets}

%{\color{blue} Opise lahko skrajšamo}

\noindent\textbf{SBVPI.} 
The first real-world dataset of the competition, SBVPI, consists of \num{1840} high-resolution RGB images from $55$ Caucasian subjects (29 females and 26 males) captured using a Canon EOS 60D DSLR camera with a macro lens under controlled conditions. 
The images contain four gaze directions (straight, left, right, and up) of each eye. $4$ samples per eye and gaze direction were acquired at varying distances and camera positions. Subjects span an age range of 15–80 years and exhibit diverse eye colors. 
%Captured images were cropped to retain only the ocular region of interest and resized to a consistent resolution of $3000 \times 1700$ pixels. 
All images have manually annotated segmentation masks of the sclera and periocular regions, while more detailed masks, which also include the iris and pupil are available for $100$ images. 
% Due to its quality and consistency, the dataset is particularly suitable for fine-grained sclera segmentation under controlled conditions. 

% {\color{blue} The dataset features extensive manual annotations, including pixel-level segmentation masks for the sclera and periocular region for all images, and additional detailed markups—such as iris, pupil, scleral vasculature, eyelashes, and canthus—for a subset of 100 images. The quality and consistency of the dataset makes it well-suited for evaluating fine-grained sclera segmentation under controlled yet subtly variable conditions.}

\vspace{1.5mm}\noindent\textbf{SMD+SLD.} The second real-world dataset, SMD+SLD, combines the Sclera Mobile Dataset~\cite{das2017towards} and the Sclera Liveness Dataset (SLD)~\cite{vitek2022exploring}, comprising $381$ RGB images of $25$ subjects from SMD and $108$ images of $27$ subjects from SLD. The images were captured by different mobile phones with an 8-mega pixel rear camera and under various acquisition conditions to increase data variety, resulting in blurry images, images with blinking eyes, and images taken at different times of the day under different lighting conditions. Consequently, the dataset enables evaluation of sclera segmentation models under non-ideal or challenging conditions. 
% The original dataset includes images with a resolution of $3264 \times 2448$ pixels, which we resize to $400 \times 300$ for the purposes of the competition. 
The dataset also contains manually generated ground truth segmentation masks for the sclera region.

\vspace{1.5mm}\noindent\textbf{MOBIUS.} The third real-world dataset of the competition, MOBIUS, was designed specifically with mobile ocular biometrics in mind.~In total, the dataset features \num{16717} high-resolution RGB images of both eyes from $100$ male and female subjects of Caucasian origin. The images were captured using three commercial mobile devices (Sony Xperia Z5 Compact, Apple iPhone 6s, and Xiaomi Pocophone F1) under varying gaze directions (straight, left, right, up) and lighting conditions (natural daylight, indoor light, low light), resulting in high image variety. For the purposes of SSBC 2025 only the part of the dataset with manually annotated segmentation masks was utilized, consisting of $3542$ images from $35$ subjects with annotated sclera, iris, and pupil regions. However, only the sclera region was considered during the evaluation process of SSBC 2025.

\subsection{Synthetic Datasets}

As SSBC 2025 focuses on the development of segmentation models with privacy-preserving synthetic data, we construct two large-scale synthetic datasets, SynCROI and SynMOBIUS, that are used for training and evaluation purposes, following the approach outlined in Figure~\ref{fig:bioculargan_framework}. % and presented in detail below. 

\vspace{1.5mm}\noindent\textbf{Synthetic-Data Generation.}~To generate synthetic data\-sets, we rely on BiOcularGAN~\cite{tomavsevic2022bioculargan}, a recent deep generative framework for creating bimodal ocular images with corresponding (synthetic, ground truth) segmentation masks.~BiOcularGAN~\cite{tomavsevic2022bioculargan} extends the StyleGAN2~\cite{karras2020analyzing} approach and achieves better image quality, by utilizing a dual-branch synthesis network, which creates aligned visible and near-infrared (NIR) ocular images based on the noise-based style information provided by the mapping network, along with two discriminator networks, one for each light spectrum, which form a joint training objective. 
In addition, BiOcularGAN relies on a separate ensemble pixel classifier to interpret latent features of the synthesis network for accurate mask generation of synthetic images~\cite{zhang2021datasetgan}. Crucially, the model also employs the adaptive discriminator augmentation (ADA)  to enable stable training even in low-data regimes~\cite{karras2020training}. We use only the generated RGB images for the competition and discard the generated NIR images.

The specific BiOcularGAN~\cite{tomavsevic2022bioculargan} implementation we employ consists of an $8$-layer mapping network, a synthesis network with $7$ synthesis blocks that outputs VIS-NIR data at resolutions from $4\times 4$ to $256\times 256$, and two discriminators, each with $7$ downsampling blocks.
We train the model in batches of $16$ with a learning rate of $0.0025$ and the Adam optimizer~\cite{kingma2014adam} with $\beta_{1} = 0$, $\beta_{2} = 0.99$, and $\epsilon=10^{-8}$. The ensemble pixel classifier, used to generate ground truth masks for the synthetic data, consists of $10$ multi-layer perceptrons trained after the initial generative model on randomly sampled pixels of manually annotated synthetic images. Training uses the cross-entropy loss and the Adam optimizer~\cite{kingma2014adam} with a learning rate of $10^{-3}$ in batches of $64$, which stops after $50$ steps with no improvement beyond the third epoch~\cite{zhang2021datasetgan}. With BiOcularGAN~\cite{tomavsevic2022bioculargan}, we generate the following synthetic datasets for SSBC 2025:%\vspace{-2mm}

\begin{figure}[tb!]
    \centering
    \includegraphics[width=0.79\linewidth]{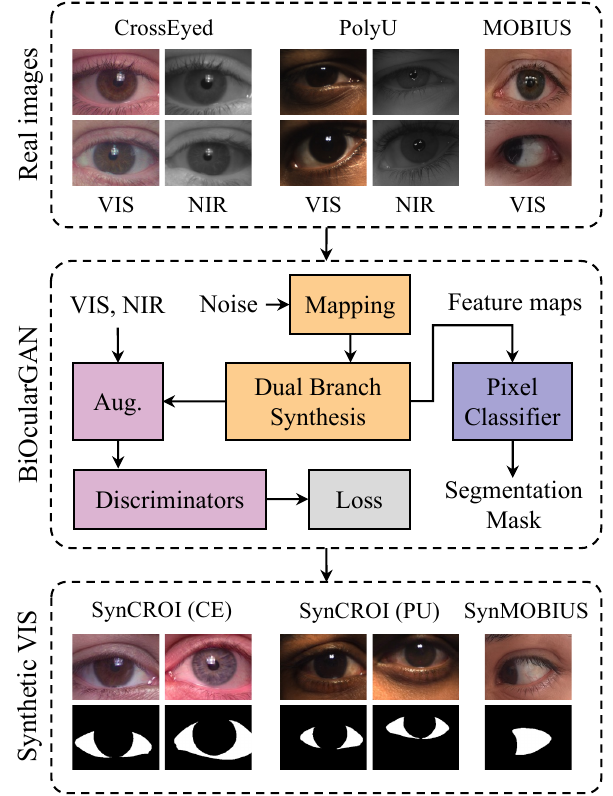}
    \caption{%{\color{blue} Dodamo če bo prostor.} 
    \textbf{Overview of data synthesis process}. To generate synthetic datasets for SSBC 2025 we train separate instances of  BiOcularGAN~\cite{tomavsevic2022bioculargan} on three real-world datasets (Cross-Eyed~\cite{sequeira2017cross}, PolyU Cross-Spectral Iris, and MOBIUS~\cite{ssbc2020}).}
    \label{fig:bioculargan_framework}\vspace{-3mm}
\end{figure}

\begin{itemize}[noitemsep,leftmargin=*]
    \item \textbf{SynCROI.} The first synthetic dataset, SynCROI, comprises \num{11000} synthetic ocular images, each accompanied by a segmentation mask of the sclera, iris, and pupil region. The dataset is divided into two large subsets with subjects of different origin, each generated with a separate instance of  BiOcularGAN~\cite{tomavsevic2022bioculargan}, trained on real-world visible (VIS) and near-infrared (NIR) image pairs.
The first subset, \textbf{SynCROI (CE)}, is produced by training BiOcularGAN~\cite{tomavsevic2022bioculargan} on the CrossEyed (CE) dataset, which includes \num{3840} VIS-NIR image pairs of $120$ Caucasian subjects with diverse eye colours. Differently, the second subset, \textbf{SynCROI (PU)}, is based on the Hong Kong Polytechnic University (PolyU) Cross-Spectral Iris Image Database, which features \num{12540} VIS-NIR image pairs of $209$ Asian subjects. 
The models are trained to convergence for $1120$ and $1600$ thousand images, respectively.  Once trained, $8$ samples are generated with each model and manually annotated for training the pixel classifiers. Afterwards, random sampling of the noise input is used with each model to produce $5500$ aligned VIS and NIR images of $256 \times 256$ pixels with corresponding segmentation masks. 
For the competition, only VIS images and their masks are considered and are resized to $400 \times 300$ to retain the aspect ratio of real-world training data. \vspace{1mm}
%Combining the two subsets ensures better demographic diversity, suitable for training more robust segmentation models with better generalization capabilities. 
\item \textbf{SynMOBIUS.} The second synthetic dataset, SynMOBIUS, is derived from the MOBIUS~\cite{ssbc2020} dataset and is used specifically for evaluation in SSBC 2025. Differently from the previous setup, the BiOcularGAN~\cite{tomavsevic2022bioculargan} model is first adapted to accommodate training on single (VIS) spectrum images by removing the second synthesis branch and its associated discriminator. The modified model is trained on MOBIUS~\cite{ssbc2020} to convergence over $2240$ thousand images. To then train the pixel classifiers, we utilize $45$ manual annotations from MOBIUS~\cite{ssbc2020}, by projecting their corresponding images to the latent space of the generative model~\cite{karras2020analyzing}, along with $5$ manually annotated synthetic samples that contain eye occlusions.
After training, the model is used to generate $5000$ images-mask pairs, which are then resized from $256\times 256$ to $400 \times 300$. \vspace{-5mm}
\end{itemize}

% and enables the evaluation of segmentation performance across different ocular appearances. 

% (trained on VIS NIR pairs for better image and semantic quality 

% Resized to $400 \times 300$... to retain the original aspect ratio of the real-world training data.

% We train all models for $2500 \ kimgs$ or until training diverges, due to the low amount of training data...

% {\color{blue}

% SSBC Synthetic - one half based on (CrossEyed) , one-half based on  The Hong Kong Polytechnic University (PolyU) Cross-Spectral Iris Images Database ...  (50-50 Caucasian / Asian subjects ... generated with BiOcularGAN (trained on VIS NIR pairs for better image and semantic quality ) from Paper ... 

% Two separate models were trained ... , to avoid issues with bimodal distributions and enable stable training ... ...
% The SMG ... were then ... trained on .... 
% and 5500 images were produced for each ... via random sampling of the noise input $z$.
%... TODO parameters? .... number of epochs ... until converng

% }
\begin{figure}[t!]
    \centering
    \includegraphics[width=0.9\linewidth]{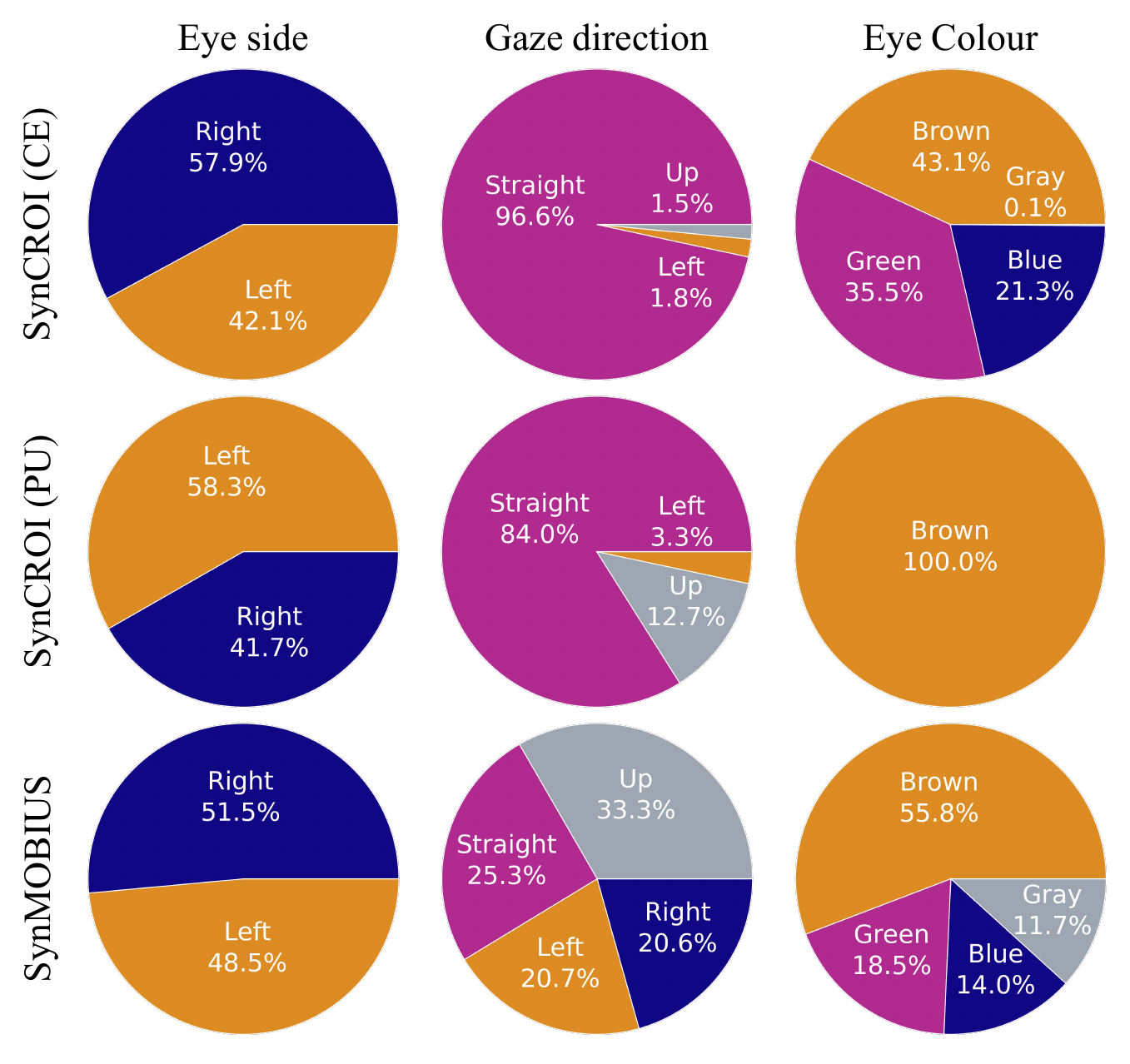}
    \caption{%{\color{blue}Lahko tudi kaj spremenimo, da bo več prostora.}  
    \textbf{Distributions of synthetic samples} over different characteristics: eye side (left, right), gaze direction (straight, left, right, up), and eye color (brown, blue, gray, green). Values are estimated with a ResNet101-based~\cite{he2016deep} classifier trained on annotations of MOBIUS and SBVPI.\vspace{-2mm}}
    \label{fig:synthetic_data_statistics}
\end{figure}

\noindent\textbf{Characteristics of Synthetic Data.}~To obtain better insight into the generated synthetic datasets, we utilized a ResNet101-based~\cite{he2016deep} classifier to determine the visual characteristics of each synthetic image, including the eye side (left or right), the gaze direction (straight, left, right, or up) and the eye colour (brown, blue, gray, or green). 
For use on the SynCROI dataset, we trained the classifier on the combined data of SBVPI and MOBIUS, which include the required annotations. Training was performed on a $9:1$ data split, over $20$ epochs in batches of $32$ and the Adam optimizer~\cite{kingma2014adam}, with an initial learning rate of $10^{-4}$ that was reduced by a factor of $10$ if no improvements were observed in $4$ epochs. Conversely, the classifier for SynMOBIUS was trained solely on the MOBIUS dataset to minimize the domain gap, with the same parameters. The distributions of the predicted characteristics for each dataset are presented in Figure~\ref{fig:synthetic_data_statistics}.~Note that both subsets of SynCROI mainly contain images with a straight gaze direction and that SynCROI (PU) only contains brown eyes, due to prevalence of these characteristics in the training data of the generative model.

\section{Benchmarking Methodology}

%%%%%%%%%%%%%%%%%%%%%%%%%%%%%%%%
\subsection{Competition Protocol}
\begin{figure}
    \centering
        \includegraphics[width=.33\columnwidth]{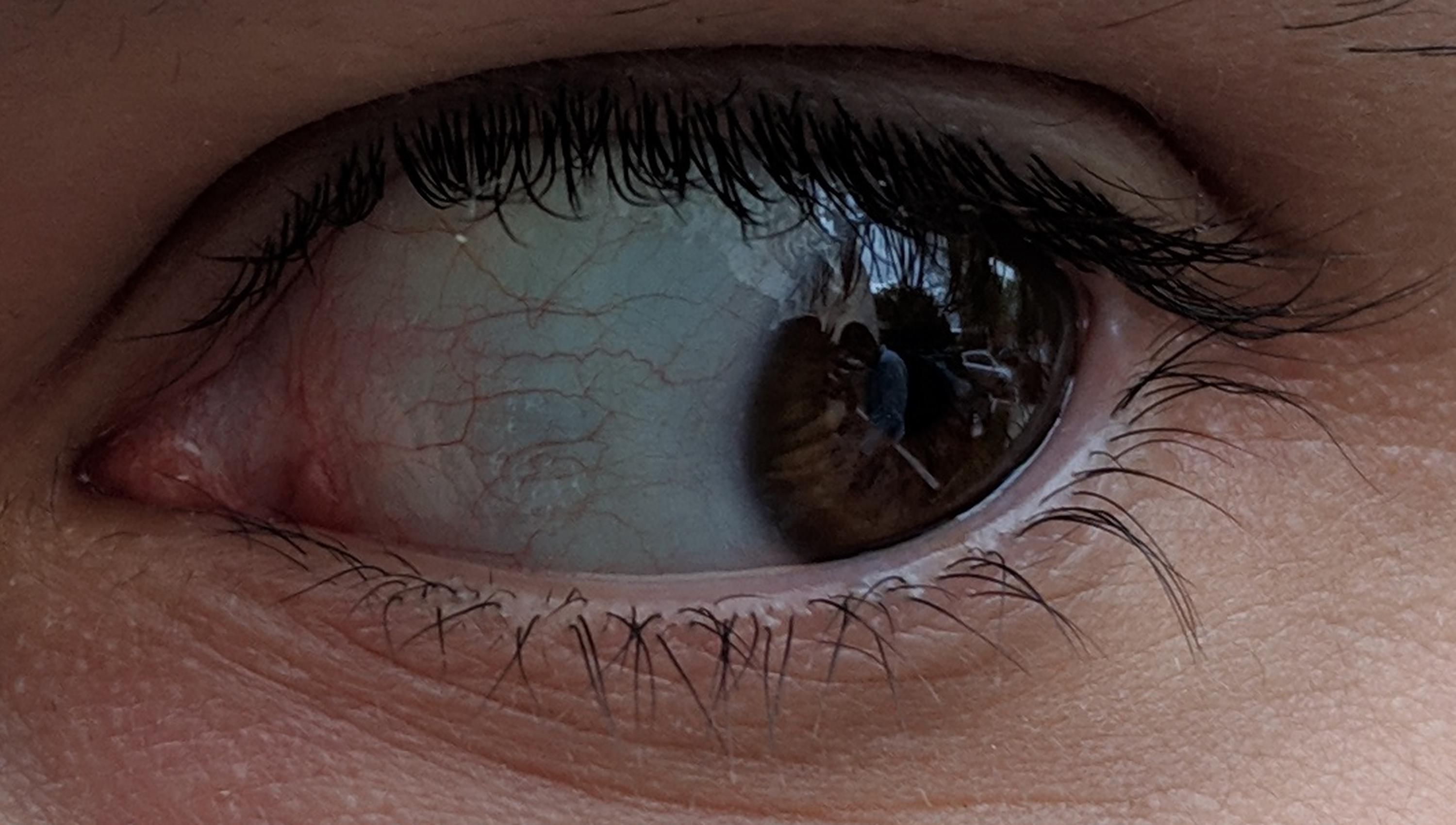}\hfill
        \includegraphics[width=.33\columnwidth]{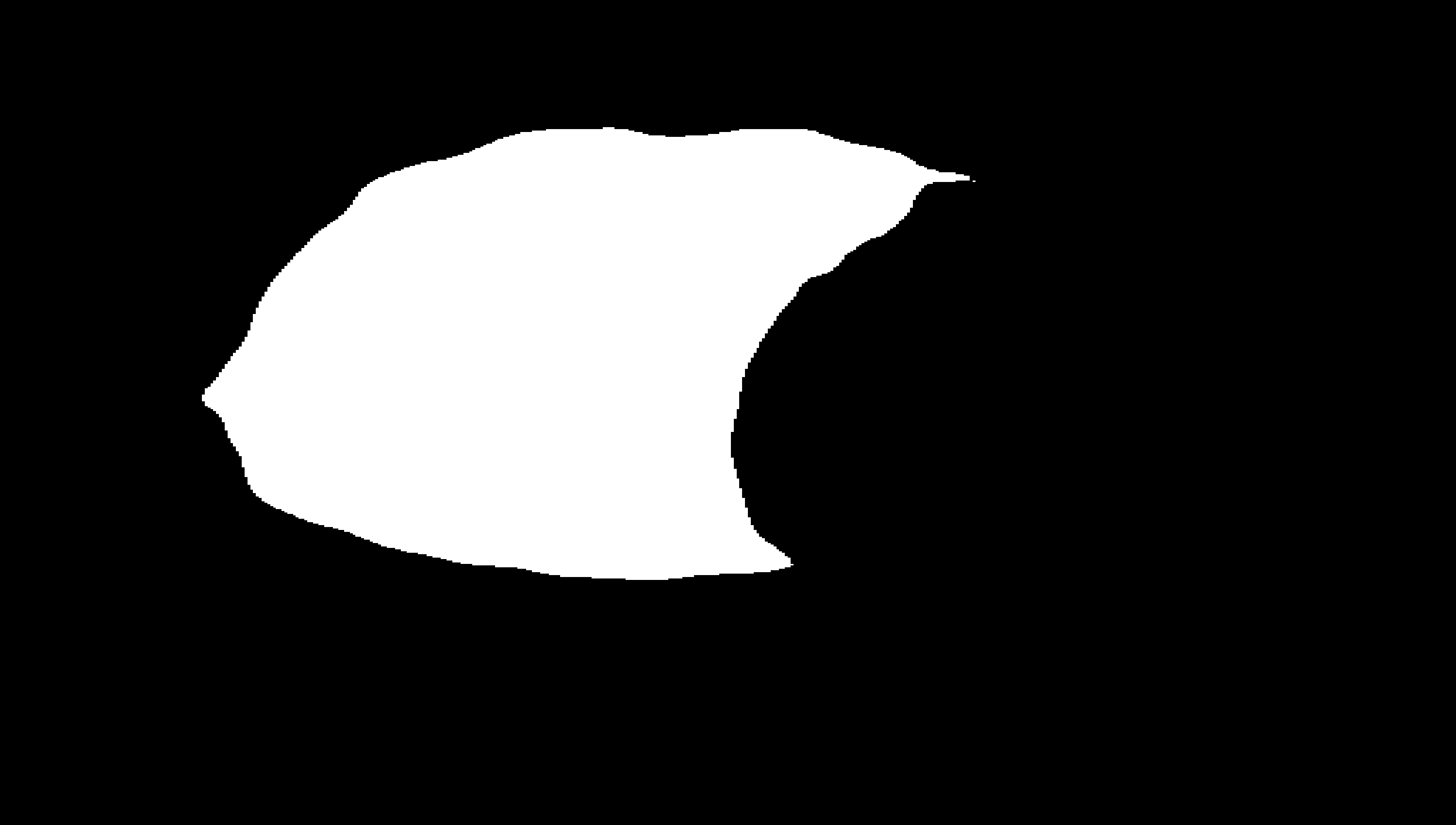}\hfill
        \includegraphics[width=.33\columnwidth]{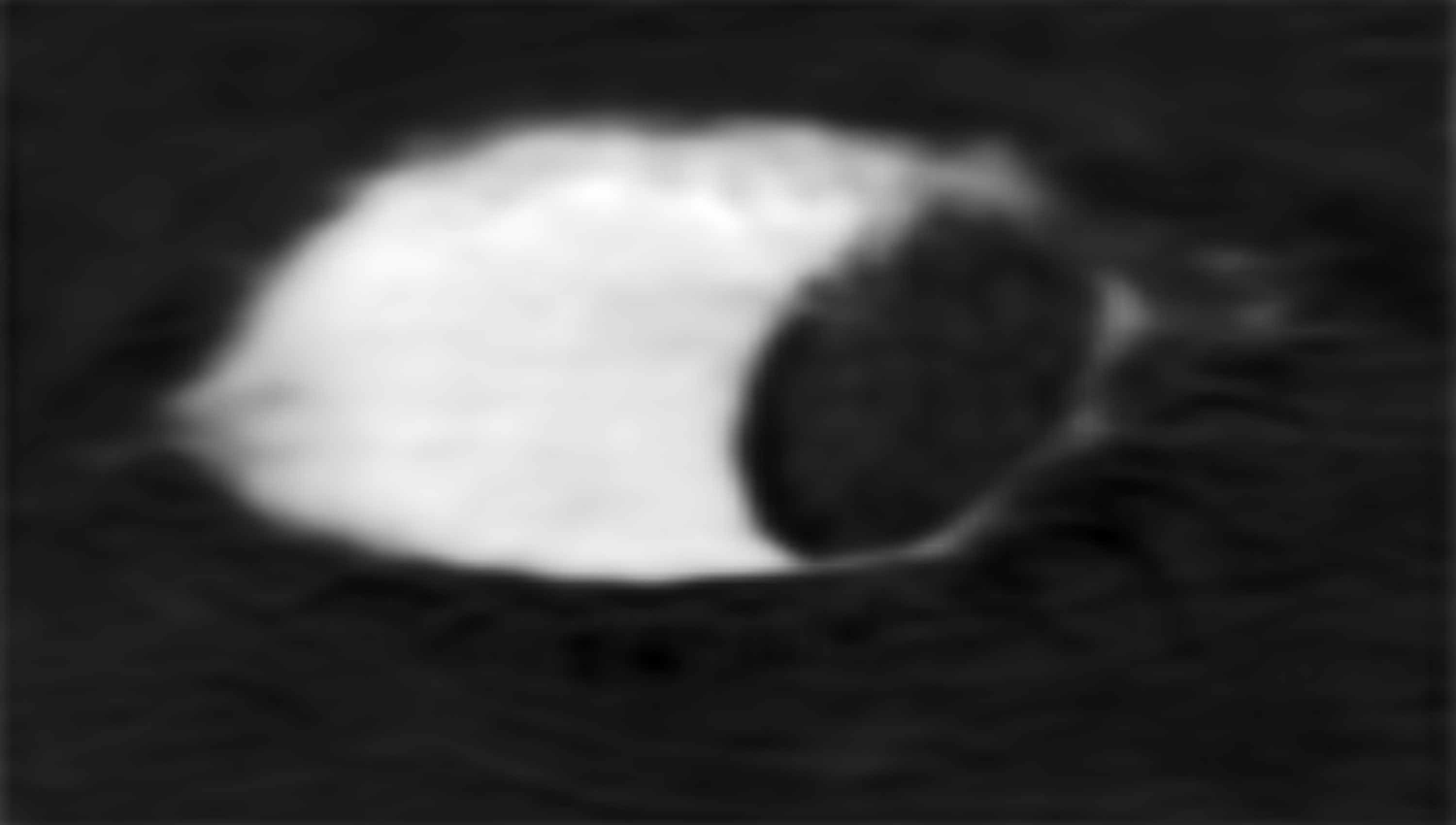}\hspace{-1mm}
    \caption{\textbf{Illustration of results to be submitted} (from left to right): original image, generated binary segmentation mask, probabilistic (grey-scale) segmentation prediction.}
    \label{fig:result_submission}
\end{figure}

SSBC 2025 entailed two separate phases.~As part of the first phase, participants were provided with real-world and synthetic training data in the form of ocular images and ground truth segmentation masks from the SBVPI and the SynCROI dataset.~The datasets were split into training and validation subsets, however, the participants were not required to use the preassigned splits.~To familiarize participants with the final submission format, a small sample of testing data was also released, comprising images of a single identity from MOBIUS and SMD+SLD, along with $10$ random samples from SynMOBIUS. These sequestered test datasets  were provided to the participants $9$ weeks after the start of the competition, without ground truth segmentation masks. The participants then only had a single week to submit their segmentation predictions, in order to limit potential fine-tuning on test data. Participants were required to deliver two types of outputs for each of the \num{14772} images from MOBIUS, SMD+SLD, and SynMOBIUS, including:  \textit{(i)} a \textbf{binary segmentation mask}, where non-zero pixels denote the sclera region and zero-valued pixels denote the background, and \textit{(ii)} \textbf{grayscale probability maps}, where pixel intensities represent the confidence of that pixel belonging to the sclera region. An example of expected output formats is provided in Figure~\ref{fig:result_submission}. The submitted binary masks served as the basis for performance ranking of participating models, while the probability maps were used for obtaining deeper insight into model behavior and enabled the derivation of detailed performance curves. 

The evaluation was performed in two distinct tracks, reflecting the focus of SSBC 2025 on the privacy-preserving use of synthetic biometric data in model development and evaluation.~The \emph{(i)} \textbf{Synthetic track} focused on models trained on synthetic data of the SynCROI dataset only, studying how such models perform on (different, sequestered) synthetic and real-world evaluation data.~The \emph{(ii)} \textbf{Mixed track}, on the other hand, allowed participants to train models on a mix of synthetic SynCROI and real-world SBVPI data, with the exact proportion being left to their discretion. As before, the trained models were evaluated on both synthetic and real-world evaluation data.

\subsection{Performance Measures}

The overall performance of the segmentation models was evaluated both through the submitted binary segmentation masks and the probabilistic predictions. For the binary masks, standard segmentation metrics were derived from true positives $TP$ (i.e. correctly detected sclera pixels), false positives $FP$ (i.e. background pixels incorrectly detected as sclera) and false negatives $FN$ (i.e. sclera pixels incorrectly detected as background), as follows: 
\begin{itemize}
    \item  \textbf{Precision}, which measures the proportion of correctly predicted sclera pixels relative to the total number of pixels predicted as sclera, i.e., $\frac{TP}{TP + FP}$~\cite{rot2018deep,lozej2018end,lozej2019influence}. %\vspace{-2mm}
    
    \item \textbf{Recall}, which quantifies the proportion of true sclera pixels that were successfully identified by the model, computed as $\frac{TP}{TP + FN}$~\cite{emervsivc2017pixel,rot2018deep,lozej2018end,lozej2019influence}. \vspace{-2mm}

    \item $\boldsymbol{F_1}$ \textbf{score}, which represents the harmonic mean of precision and recall, defined as $2 \cdot \frac{\text{Precision} \cdot \text{Recall}}{\text{Precision} + \text{Recall}}$. This aggregated metric balances the trade-off between precision and recall and serves as the main criterion for ranking participating models. \vspace{-2mm}

    \item \textbf{Intersection over union (IoU)}, or the Jaccard index, captures the overlap between the predicted and the ground truth sclera regions, normalized by their union, calculated as $\frac{TP}{TP + FP + FN}$.
    
\end{itemize}
% In the above equations $TP$ denotes {\em true positives}, i.e., the number of correctly retrieved sclera pixels, $FP$ denotes {\em false positives}, i.e., the number of background pixels incorrectly retrieved as sclera pixels, and $FN$ denotes {\em false negatives}, i.e., the number of sclera pixels incorrectly retrieved as background pixels.

To provide a more comprehensive assessment beyond binary classifications, the grayscale (probabilistic) segmentation maps were also used to produce \textbf{precision-recall curves}~\cite{powers2011evaluation,saito2015precision}. From the curves, the optimal $F_1$ score ($F_1^{opt}$) and the \textbf{Area Under the Curve} (AUC)~\cite{boyd2013area} can be computed, offering a more nuanced comparison of models across varying confidence thresholds.

\section{Summary of Submitted Approaches}

A total of $9$ teams submitted their entries to SSBC 2025. %All the submissions relied on deep learning, which is in line with the goals of our competition, but also reflects the trend the field has been moving towards \cite{ssbc2023,vitek2022exploring}. 
\Cref{tab:methods} presents a summary of the submissions, while a brief description of each of the entries is provided below. %We also note how the teams addressed the issues arising from the synthetic data training (Synthetic track in \cref{sec:experiments}) and the mixing of synthetic and real-world data (Mixed track in \cref{sec:experiments}).

\begin{table}
\renewcommand{\arraystretch}{1.1}
   \caption{\textbf{List of submitted entries to SSBC 2025} and their respective institutions. The abbreviations used for the models in this table correspond to the ones used in the experimental section. %\textcolor{blue}{Tukaj lahko damo še kakšne lastnosti, ki bi bile zanimive, če imamo prostor za 2-stolpčno tabelo in če bo dovolj časa.}
   }
  \label{tab:methods}
    \centering
    \resizebox{\columnwidth}{!}{%
    \begin{tabular}{llr}
        %\hline%\hline
        \toprule
        Team & Model Acronym & Base$^\dagger$ \\\midrule
        Indian Institute of Technology Mandi (IIT-M) & SEG-U-Sclera & SAM \\
        Beijing University of Civil Engineering and Architecture (BUCEA) & SAM2-UNet & SAM/UN \\
        Warsaw University of Technology (WUT) & AEOS & SAM/SF \\
        Khalifa University (KU) & KU-CVML & DL \\
        Ahmedabad University (AU) & ShapeGAN-DLV3+ & DL \\
        Couger Inc. & SwinDANet & ST/DN \\
        SRI International & UL-VMUNet & M/UN \\
        Idiap -- University of Applied Sciences and Arts Western Switzerland (HES-SO) & SAM-Iris & SAM \\
        Pandit Deendayal Energy University (PDPU) & UNet++\_Binary & UN \\
        %\hline\hline
        \bottomrule
        \multicolumn{3}{l}{For details on the participants from the institutions, see the author list.}\\
        \multicolumn{3}{l}{$^\dagger$Base architecture: DL -- DeepLab \cite{chen2018deeplab}, DN -- DenseNet \cite{huang2017densely}, M -- Mamba \cite{gu2023mamba}, SAM -- Segment Anything Model \cite{kirillov2023segment}}\\
        \multicolumn{3}{l}{SF -- SegFormer \cite{xie2021segformer}, ST -- Swin Transformer \cite{liu2022swin}, UN -- U-Net \cite{unet}.}
        \vspace{-4mm}
   \end{tabular}
   }
\end{table}

%\textcolor{blue}{Bi dali v appendix celotne opise pristopov, kot so nam jih poslali? Glede na to, da so tukaj precej strnjeni.} {\color{purple} Darian: Mogoče ali so uporabili 2 ali 4 classe?}

\vspace{1.2mm}\noindent\textbf{SEG-U-Sclera (IIT-M)} is a variant of SAM2 \cite{ravi2024sam} trained with uncertainty-weighted binary cross-entropy (BCE). This loss addresses the difference between real-world and synthetic data by focusing on the uncertain regions where synthetic and real-world images vary.

\vspace{1.2mm}\noindent\textbf{SAM2-UNet (BUCEA)} is another SAM2-based\cite{ravi2024sam} model, utilizing a U-Net-like architecture \cite{Xiong2024SAM2UNetSA} with a Hiera encoder \cite{pmlr-v202-ryali23a} and a U-shaped 3-block decoder. Lightweight adapters are inserted into the parameter-frozen Hiera backbone to ensure parameter-efficient fine-tuning for the sclera segmentation task.~The model is trained with weighted IoU and BCE losses and deep supervision strategies.

\vspace{1.2mm}\noindent\textbf{AEOS (WUT)}: the WUT team used a hybrid approach, with their architecture depending on the amount of training data. Their SAM2-based \cite{ravi2024sam} architecture tended to underperform on small amounts of training data, so for the Synthetic track, they relied on a SegFormer-like \cite{xie2021segformer} model. This hybrid approach resulted in better generalization, as it explicitly addressed small/large training datasets.

\vspace{1.2mm}\noindent\textbf{KU-CVML (KU)} relies on the DeepLabV3+ \cite{chen2018deeplab} architecture and the EfficientNet-B4 \cite{EfficientNet} encoder and is trained with a hybrid loss, which is partially inspired by semi-supervised training approaches and SAM \cite{kirillov2023segment}, and also addresses class imbalance and contour accuracy.

\vspace{1.2mm}\noindent\textbf{ShapeGAN-DLV3+ (AU)} is an extension of DeepLabV3+ \cite{chen2018deeplab} that consists of several sclera-segmentation-specific modules, i.e.:~$(i)$ a combination of CoordConv \cite{liu2018intriguing} and deformable stem convolution \cite{zhu2019deformable} to provide spatial awareness and adaptive receptive fields; $(ii)$ a FourierLoss \cite{erden2025fourierloss} to enforce shape fidelity; $(iii)$ an InverseFormLoss \cite{borse2021inverseform} to promote fine boundary alignment; and $(iv)$ a PatchGAN discriminator that is attached to the end of the encoder and trained in an adversarial manner, to provide a bridge crossing the synthetic-to-real domain gap.

\vspace{1.2mm}\noindent\textbf{SwinDANet (Couger)} is a hybrid encoder-decoder model. The encoder is based on the Swin Transformer V2 \cite{liu2022swin} with shifted-window self-attention, which enables the modeling of long-range dependencies and extracts multi-scale features in a hierarchical contextual representation. The decoder is DenseNet-based \cite{huang2017densely}, augmented with CSSE attention \cite{kansal2019eyenet}, emphasizing salient spatial and channel-wise information, and adaptive skip-connections that preserve spatial details from the encoder.

\vspace{1.2mm}\noindent\textbf{UL-VMUNet (SRI)} is a lightweight Mamba \cite{gu2023mamba} variant, combining the state-space Mamba model with a U-Net-like architecture \cite{wu2024ultralight} to attain compute- and parameter-efficient high-quality segmentations. To generalize better from synthetic data, hand-crafted data augmentation is performed to mimic different lighting, poses, and camera artifacts.

\vspace{1.2mm}\noindent\textbf{SAM-Iris (Idiap -- HES-SO)} is a SAM2-based \cite{ravi2024sam} model, fine-tuned with a sequential synthetic-to-real training process inspired by \cite{tremblay2018training}. The model additionally contains a prompt point in the center of the input eye image, which ensures that the model focuses specifically on the eye region.

\vspace{1.2mm}\noindent\textbf{UNet++\_Binary (PDPU)} adopts the UNet++ \cite{zhou2018unet++} architecture, using a ResNet-style \cite{resnet} encoder. It is fine-tuned for the task of sclera segmentation on images normalized using ImageNet \cite{imagenet_cvpr09} statistics.

\begin{figure}[t]
\centering
    \begin{subfigure}{\linewidth}
        \includegraphics[width=\textwidth]{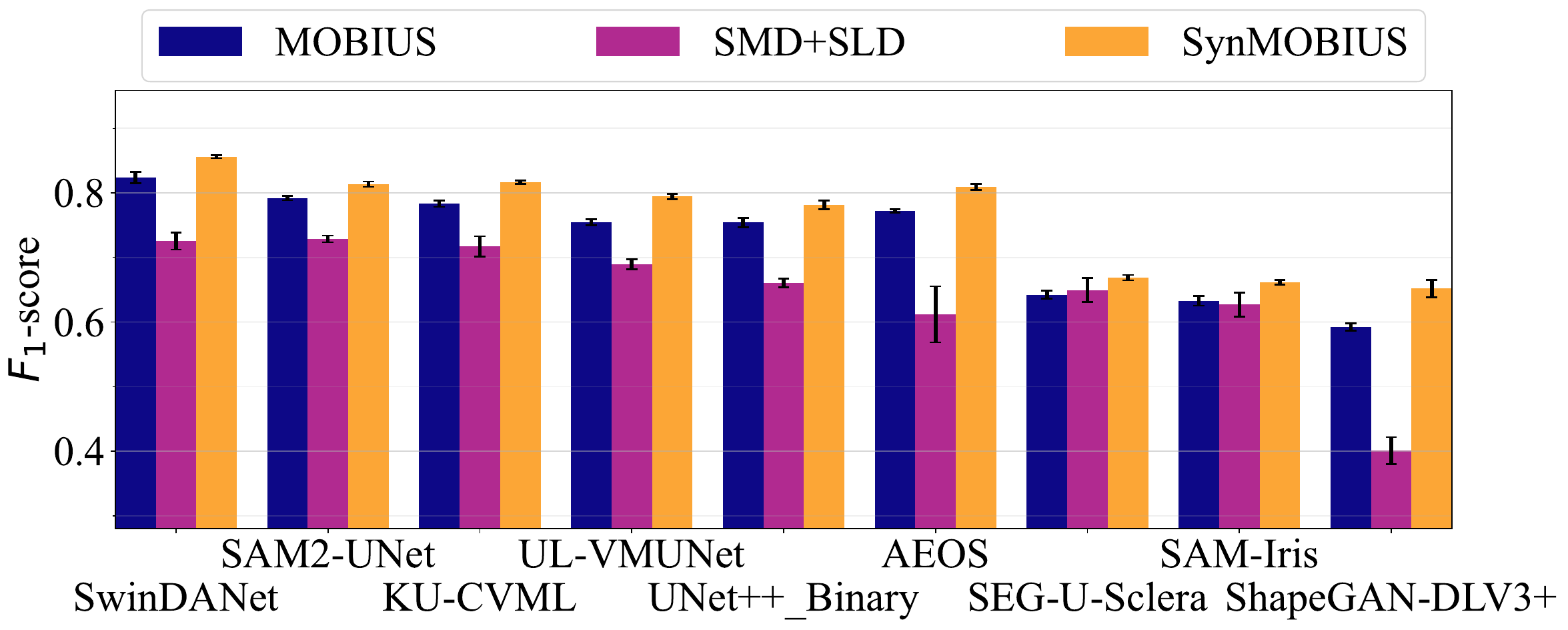}
        \caption{Synthetic track}
        \label{fig:bar_synth}
    \end{subfigure}\\
    \begin{subfigure}{\linewidth}
        \includegraphics[width=\textwidth]{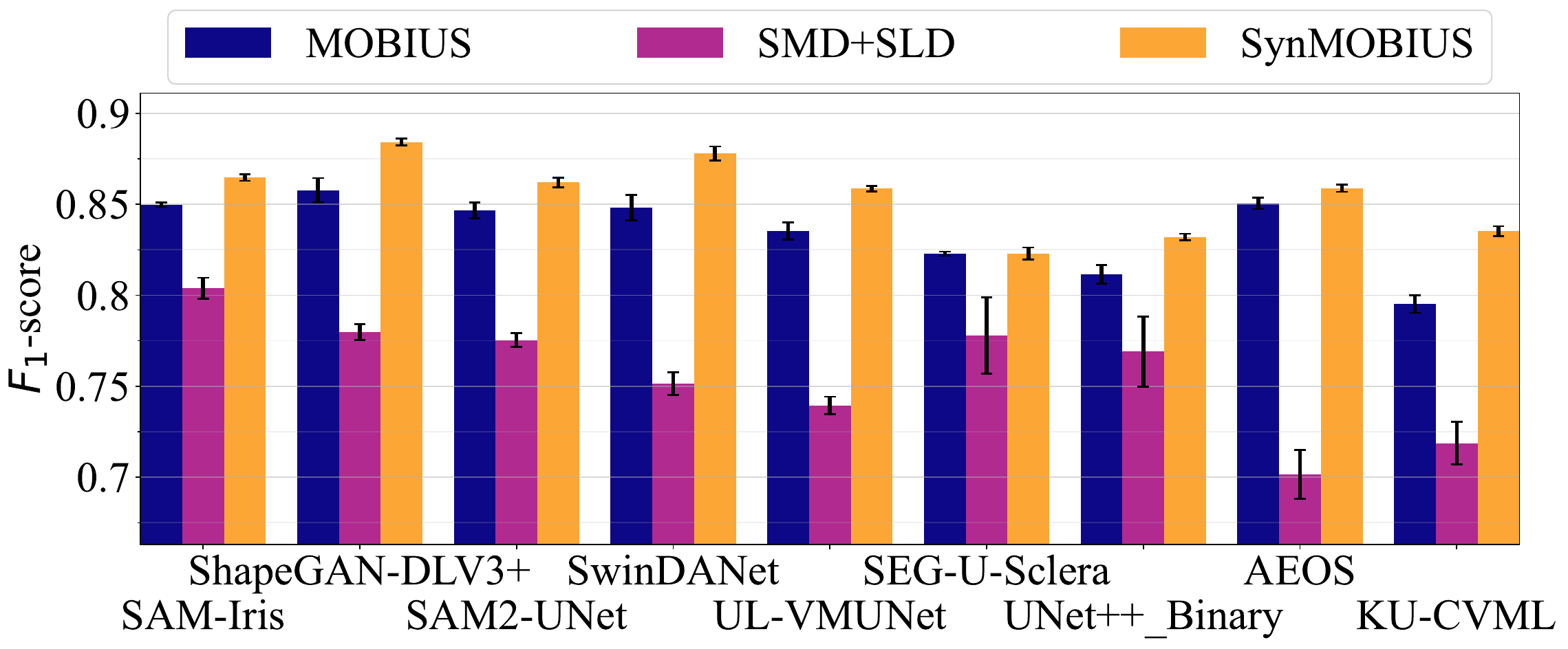}
        \caption{Mixed track}
        \label{fig:bar_mixed}
    \end{subfigure}
    \caption{\textbf{Performance comparison} of the submissions trained on \subref{fig:bar_synth} synthetic and \subref{fig:bar_mixed} mixed training data, in terms of the $F_1$ score achieved over all the images in the evaluation datasets.}\vspace{-3mm}
    \label{fig:bar_overall}
\end{figure}

%\begin{table}%  HMean only
\begin{table*}%  Full
    \caption{\textbf{Comparative assessment of the models trained on synthetic data}. The submissions are ranked according to the harmonic means of the achieved $F_1$ scores over the three evaluation datasets. The $F_1$, Precision, Recall, and IoU scores were computed from the submitted binary masks. The optimal $F_1$ score on the precision-recall curve ($F_1^{opt}$) and AUC values were calculated from the probabilistic segmentation predictions. The harmonic means are also reported for the rest of the performance measures in the column next to the individual dataset results. Note that the ranking is quite consistent across performance indicators. \vspace{1mm}%\textcolor{blue}{V teh dveh tabelah bi mogoče lahko kazal samo (harmonično) povprečje pri vsakem scoru -- se takoj precej zmanjša tabela in bi šla mogoče celo v en stolpec lahko... Vsaj za F1 imamo performance po posameznih datasetih tudi v bar grafih, tako da ne zgubimo povsem te informacije. Sem pripravil, samo odkomentiraj vrstice, ki imajo \% HMEan only in zakomentiraj, kjer je \% Full.}
    }
    \label{tab:results_synth}
    \renewcommand{\arraystretch}{1.1}
    \centering
    \resizebox{0.93\linewidth}{!}{
    \begin{tabular}{rll cccccccc cccc}%  Full
    %\rowcolors{5}{white}{gray!10}%  HMean only
    %\begin{tabular}{rl cccc cc}%  HMean only
        %\hline\hline
        \toprule
        \multirow{2}{*}{Rank} & \multirow{2}{*}{Segmentation Model} & \multirow{2}{*}{Evaluation Dataset} & \multicolumn{8}{c}{From binary masks} & \multicolumn{4}{c}{From probabilistic predictions}\\\cmidrule(lr){4-11} \cmidrule(lr){12-15}%  Full
        & & & \multicolumn{2}{c}{$F_1$} & \multicolumn{2}{c}{Precision} & \multicolumn{2}{c}{Recall} & \multicolumn{2}{c}{IoU} & \multicolumn{2}{c}{$F_1^{opt}$} & \multicolumn{2}{c}{AUC}\\\hline%  Full
        %\multirow{2}{*}{Rank} & \multirow{2}{*}{Segmentation Model} & \multicolumn{4}{c}{Binary} & \multicolumn{2}{c}{Probabilistic}\\\cmidrule(lr){3-6} \cmidrule(lr){7-8}%  HMean only
        %& & $F_1$ & Precision & Recall & IoU & $F_1^{opt}$ & AUC\\\hline%  HMean only
        \multirow{3}{*}{1} & \multirow{3}{*}{SwinDANet} & MOBIUS & \num{0.824} & \multirow{3}{*}{\num{0.798}} & \num{0.825} & \multirow{3}{*}{\num{0.754}} & \num{0.842} & \multirow{3}{*}{\num{0.868}} & \num{0.722} & \multirow{3}{*}{\num{0.680}} & \num{0.850} & \multirow{3}{*}{\num{0.824}} & \num{0.887} & \multirow{3}{*}{\num{0.870}} \\
 & & SMD+SLD   & \num{0.725} & & \num{0.640} & & \num{0.866} & & \num{0.587} & & \num{0.758} & & \num{0.809} & \\
 & & SynMOBIUS &  \num{0.856} & & \num{0.830} & & \num{0.897} & & \num{0.757} & & \num{0.875} & & \num{0.921} & \\\hline
\multirow{3}{*}{2} & \multirow{3}{*}{SAM2-UNet} & MOBIUS & \num{0.792} & \multirow{3}{*}{\num{0.776}} & \num{0.792} & \multirow{3}{*}{\num{0.728}} & \num{0.822} & \multirow{3}{*}{\num{0.856}} & \num{0.677} & \multirow{3}{*}{\num{0.649}} & \num{0.821} & \multirow{3}{*}{\num{0.803}} & \num{0.843} & \multirow{3}{*}{\num{0.826}} \\
 & & SMD+SLD   & \num{0.729} & & \num{0.636} & & \num{0.872} & & \num{0.584} & & \num{0.756} & & \num{0.779} & \\
 & & SynMOBIUS &  \num{0.813} & & \num{0.777} & & \num{0.876} & & \num{0.698} & & \num{0.837} & & \num{0.861} & \\\hline
\multirow{3}{*}{3} & \multirow{3}{*}{KU-CVML} & MOBIUS & \num{0.783} & \multirow{3}{*}{\num{0.770}} & \num{0.711} & \multirow{3}{*}{\num{0.676}} & \num{0.914} & \multirow{3}{*}{\num{0.925}} & \num{0.667} & \multirow{3}{*}{\num{0.641}} & - & \multirow{3}{*}{-} & - & \multirow{3}{*}{-} \\
 & & SMD+SLD   & \num{0.717} & & \num{0.596} & & \num{0.926} & & \num{0.571} & & - & & - & \\
 & & SynMOBIUS &  \num{0.816} & & \num{0.738} & & \num{0.937} & & \num{0.700} & & - & & - & \\\hline
\multirow{3}{*}{4} & \multirow{3}{*}{UL-VMUNet} & MOBIUS & \num{0.755} & \multirow{3}{*}{\num{0.744}} & \num{0.822} & \multirow{3}{*}{\num{0.728}} & \num{0.736} & \multirow{3}{*}{\num{0.793}} & \num{0.633} & \multirow{3}{*}{\num{0.609}} & \num{0.817} & \multirow{3}{*}{\num{0.791}} & \num{0.859} & \multirow{3}{*}{\num{0.836}} \\
 & & SMD+SLD   & \num{0.689} & & \num{0.600} & & \num{0.839} & & \num{0.537} & & \num{0.729} & & \num{0.772} & \\
 & & SynMOBIUS &  \num{0.794} & & \num{0.808} & & \num{0.811} & & \num{0.674} & & \num{0.836} & & \num{0.885} & \\\hline
\multirow{3}{*}{5} & \multirow{3}{*}{UNet++\_Binary} & MOBIUS & \num{0.754} & \multirow{3}{*}{\num{0.728}} & \num{0.851} & \multirow{3}{*}{\num{0.784}} & \num{0.720} & \multirow{3}{*}{\num{0.716}} & \num{0.643} & \multirow{3}{*}{\num{0.599}} & \num{0.832} & \multirow{3}{*}{\num{0.806}} & \num{0.878} & \multirow{3}{*}{\num{0.840}} \\
 & & SMD+SLD   & \num{0.660} & & \num{0.690} & & \num{0.668} & & \num{0.513} & & \num{0.746} & & \num{0.762} & \\
 & & SynMOBIUS &  \num{0.781} & & \num{0.834} & & \num{0.769} & & \num{0.666} & & \num{0.848} & & \num{0.893} & \\\hline
\multirow{3}{*}{6} & \multirow{3}{*}{AEOS} & MOBIUS & \num{0.772} & \multirow{3}{*}{\num{0.720}} & \num{0.914} & \multirow{3}{*}{\num{0.847}} & \num{0.714} & \multirow{3}{*}{\num{0.677}} & \num{0.683} & \multirow{3}{*}{\num{0.622}} & \num{0.871} & \multirow{3}{*}{\num{0.822}} & \num{0.921} & \multirow{3}{*}{\num{0.867}} \\
 & & SMD+SLD   & \num{0.612} & & \num{0.731} & & \num{0.589} & & \num{0.511} & & \num{0.731} & & \num{0.763} & \\
 & & SynMOBIUS &  \num{0.809} & & \num{0.927} & & \num{0.749} & & \num{0.712} & & \num{0.883} & & \num{0.941} & \\\hline
\multirow{3}{*}{7} & \multirow{3}{*}{SEG-U-Sclera} & MOBIUS & \num{0.642} & \multirow{3}{*}{\num{0.653}} & \num{0.655} & \multirow{3}{*}{\num{0.621}} & \num{0.669} & \multirow{3}{*}{\num{0.725}} & \num{0.512} & \multirow{3}{*}{\num{0.524}} & \num{0.716} & \multirow{3}{*}{\num{0.709}} & \num{0.642} & \multirow{3}{*}{\num{0.632}} \\
 & & SMD+SLD   & \num{0.650} & & \num{0.569} & & \num{0.782} & & \num{0.521} & & \num{0.685} & & \num{0.593} & \\
 & & SynMOBIUS &  \num{0.669} & & \num{0.644} & & \num{0.733} & & \num{0.541} & & \num{0.728} & & \num{0.666} & \\\hline
\multirow{3}{*}{8} & \multirow{3}{*}{SAM-Iris} & MOBIUS & \num{0.633} & \multirow{3}{*}{\num{0.640}} & \num{0.639} & \multirow{3}{*}{\num{0.619}} & \num{0.657} & \multirow{3}{*}{\num{0.693}} & \num{0.487} & \multirow{3}{*}{\num{0.497}} & \num{0.675} & \multirow{3}{*}{\num{0.677}} & \num{0.683} & \multirow{3}{*}{\num{0.672}} \\
 & & SMD+SLD   & \num{0.627} & & \num{0.574} & & \num{0.716} & & \num{0.489} & & \num{0.658} & & \num{0.631} & \\
 & & SynMOBIUS &  \num{0.662} & & \num{0.648} & & \num{0.710} & & \num{0.516} & & \num{0.700} & & \num{0.707} & \\\hline
\multirow{3}{*}{9} & \multirow{3}{*}{ShapeGAN-DLV3+} & MOBIUS & \num{0.592} & \multirow{3}{*}{\num{0.525}} & \num{0.743} & \multirow{3}{*}{\num{0.612}} & \num{0.539} & \multirow{3}{*}{\num{0.545}} & \num{0.482} & \multirow{3}{*}{\num{0.396}} & \num{0.729} & \multirow{3}{*}{\num{0.629}} & \num{0.779} & \multirow{3}{*}{\num{0.661}} \\
 & & SMD+SLD   & \num{0.401} & & \num{0.430} & & \num{0.517} & & \num{0.275} & & \num{0.483} & & \num{0.492} & \\
 & & SynMOBIUS &  \num{0.652} & & \num{0.813} & & \num{0.583} & & \num{0.534} & & \num{0.755} & & \num{0.817} &
 \\%  Full
        %\input{Synthetic - HMean} \\%  HMean only
        %\hline\hline
        \bottomrule
        \multicolumn{15}{l}{The probabilistic results for KU-CVML are not reported due to issues in the submission.}%  Full
        %\hiderowcolors%  HMean only
        %\multicolumn{8}{l}{The probabilistic results for KU-CVML are not reported due to issues in the submission.}%  HMean only
    \end{tabular}%
    }\vspace{-2mm}
%\end{table}%  HMean only
\end{table*}%  Full

%%%%%%%%%%%%%%%%%%%%%%%%%%%%%%%%%%%%%%%%%%%%%%%%%%%%%%%%%%%%%%%%%%%%%%%%%%%%%%%%%%%%%
\section{Benchmarking Results} % with analysis and discussion}
\label{sec:experiments}
%%%%%%%%%%%%%%%%%%%%%%%%%%%%%%%%%%%%%%%%%%%%%%%%%%%%%%%%%%%%%%%%%%%%%%%%%%%%%%%%%%%%%

In this section, we present the results of SSBC 2025, both for the synthetic and the mixed track.~Additionally, we perform a detailed analysis of the performance of the submitted models across different segmentation thresholds and study the performance differentials between the corresponding models from the two competition tracks.
%The evaluation this year was performed in two distinct tracks, reflecting our focus on the privacy-preserving use of synthetic biometric data in model development and evaluation. The \emph{(i) Synthetic} track focuses on models trained on synthetic data only, studying how such models perform on other synthetic, as well as real-world evaluation data. The \emph{(ii) Mixed} track allows the models to be trained on the additional real-world data of the SBVPI dataset, again evaluating the models on both synthetic and real-world evaluation data.

\subsection{Overall Results and Performance Ranking}
\label{sec:results}

To evaluate and rank the SSBC 2025 submissions, we computed \textbf{average performance scores} over the submitted segmentation masks. The standard errors reported in this section were obtained by partitioning the test data into $5$ subject-disjoint folds and computing the corresponding standard deviation. The harmonic mean $F_1$ score computed across the three evaluation datasets was used as the criterion for model ranking. This ensured that the scoring system rewards the submissions that maintain a steady high performance across all evaluation datasets, rather than performing well on some datasets and failing on others.

%\label{sec:results_synth}
\begin{figure*}[t]
    \centering
    \includegraphics[width=.26\linewidth]{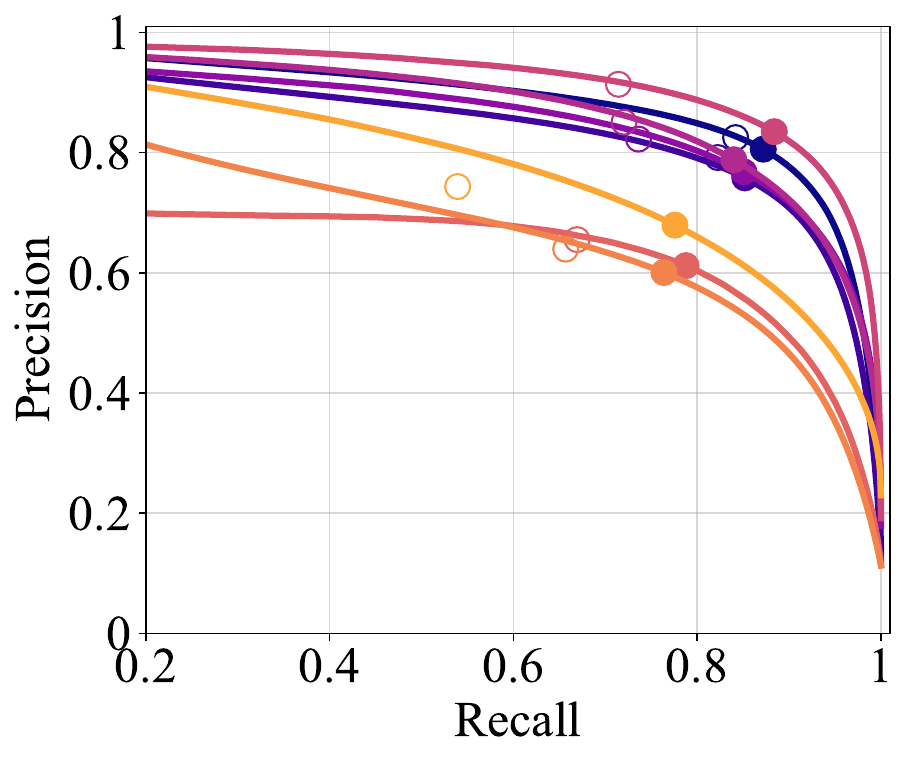}\hfill%
    \includegraphics[width=.26\linewidth]{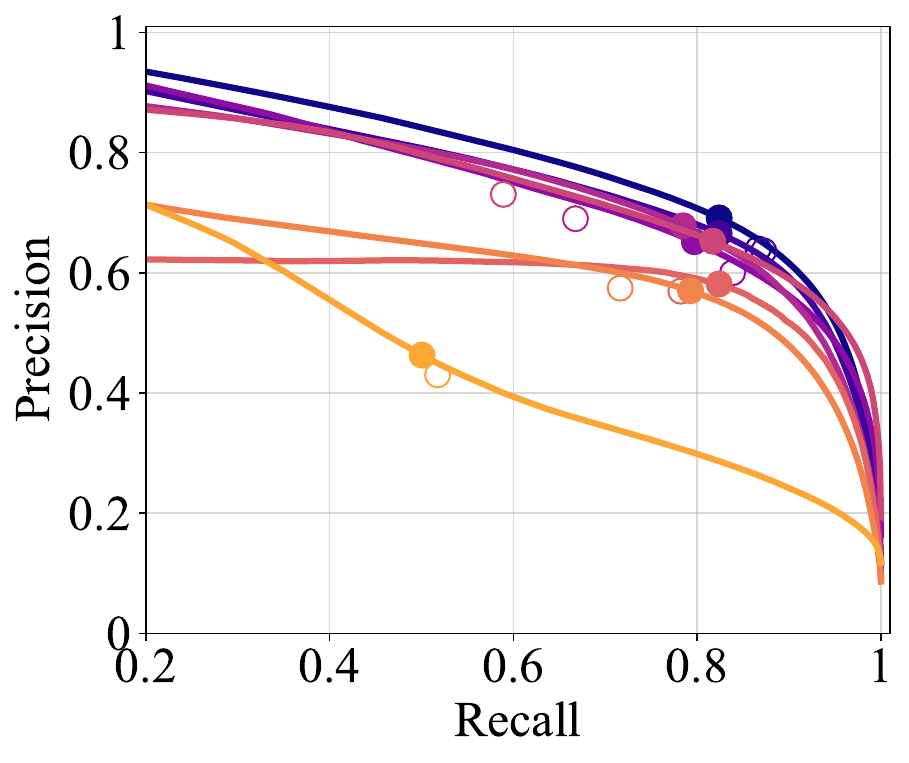}\hfill%
    \includegraphics[width=.43\linewidth]{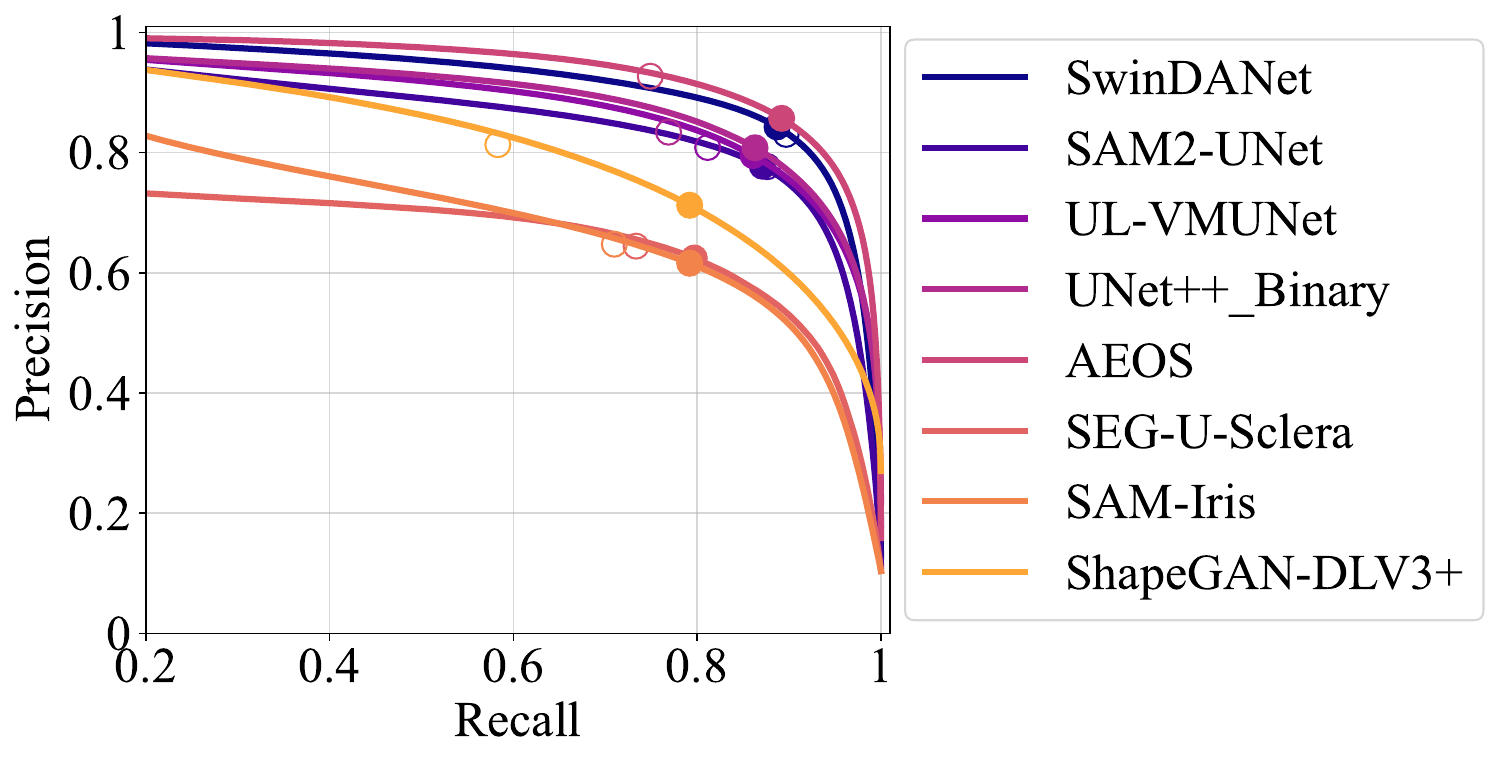}\\
    \includegraphics[width=.26\linewidth]{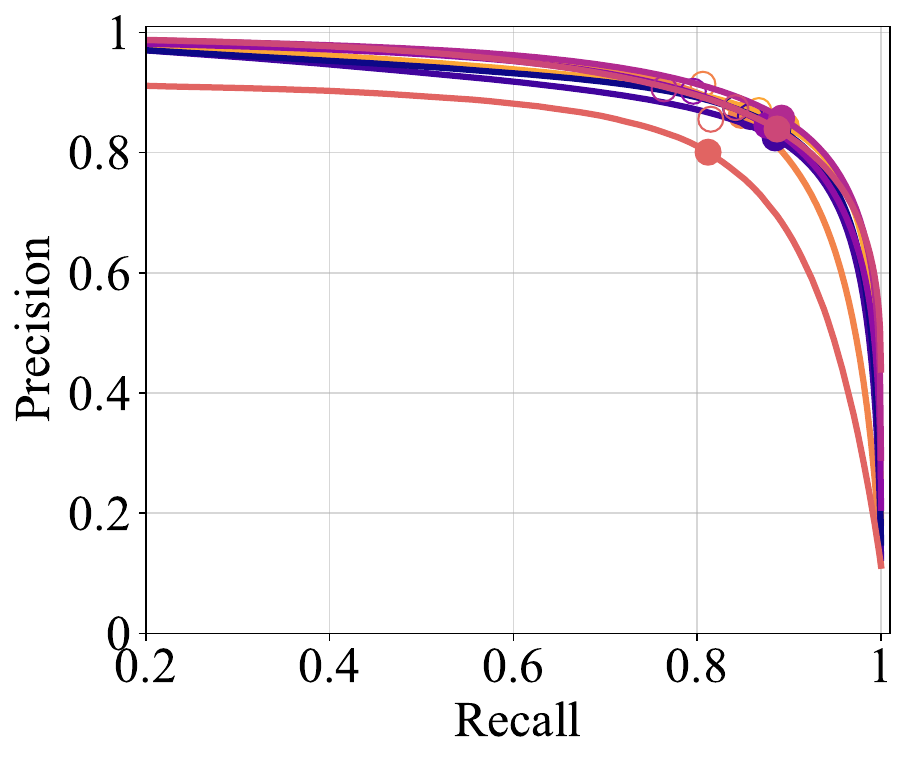}\hfill%
    \includegraphics[width=.26\linewidth]{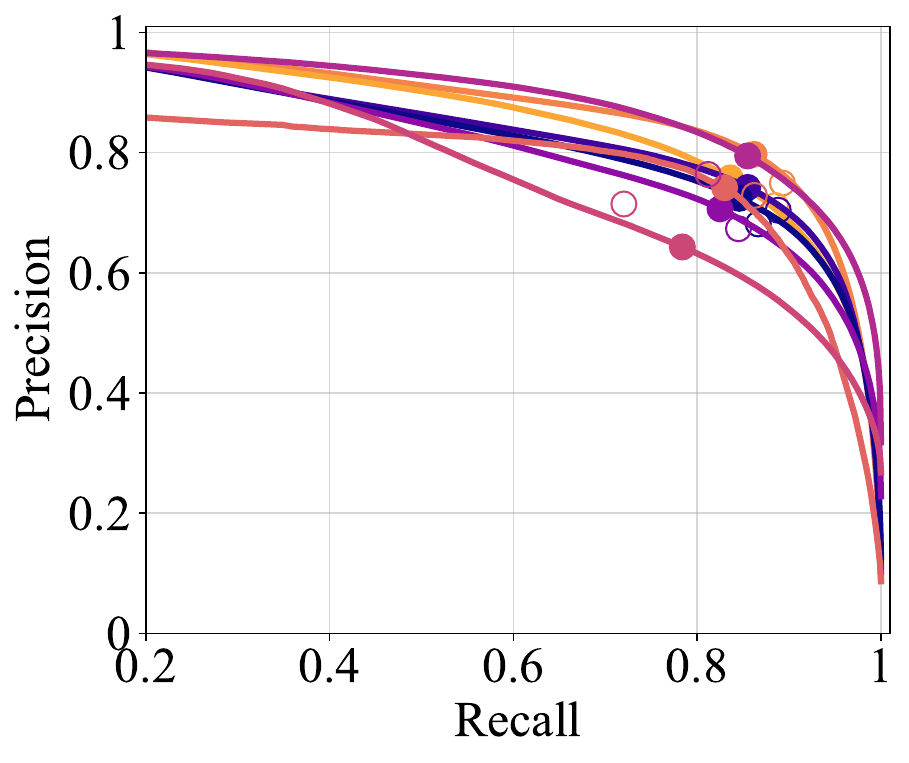}\hfill%
    \includegraphics[width=.43\linewidth]{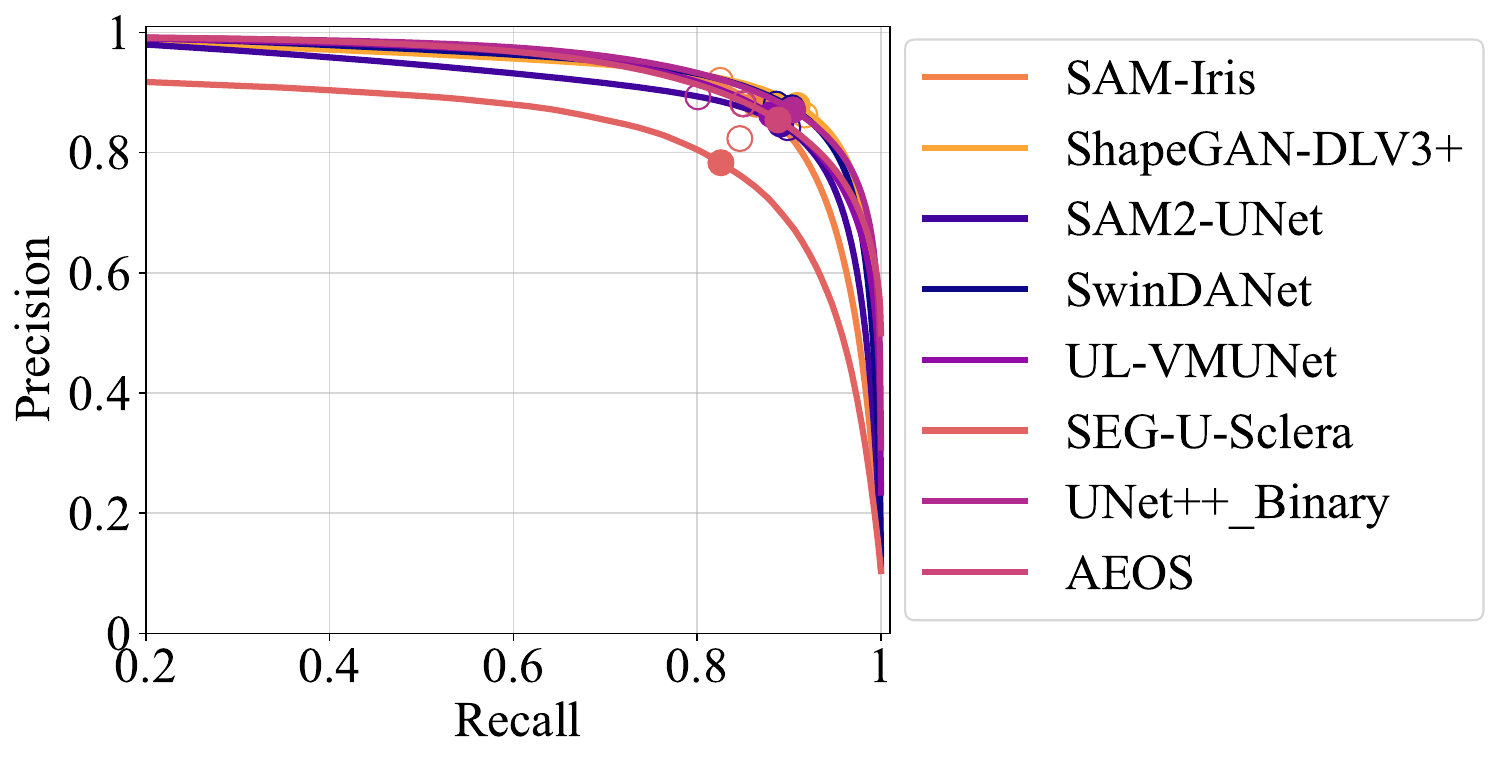}\\
    \caption{\textbf{Precision-recall curves for the submitted SSBC 2025 models}. The operating points denoted with a full circle represent the best possible $F_1$ score ($F_1^{opt}$), whereas the empty circle denotes the precision-recall point produced by the binary masks. The first row displays the results of the submissions trained on synthetic data, whereas the second row contains submissions trained in the mixed-data scenario. The columns show the results on the evaluation datasets in following order: MOBIUS, SMD+SLD and SynMOBIUS. The legends are sorted by the corresponding track's ranking. The figure is best viewed in color and zoomed in.}
    \label{fig:pr_overall}
\end{figure*}

\vspace{1.5mm}\noindent\textbf{Synthetic Track.}~Since SSBC 2025 focuses on the use of synthetically generated data in model development, the first track of the competition benchmarked models trained solely on synthetic data. The results of this track are presented in \cref{fig:bar_synth,tab:results_synth}. The winner of the Synthetic track is \textbf{SwinDANet}, which outperformed all other approaches in all the computed performance metrics. Most models, however, achieved reasonably competitive results, with all but 3 of the submissions landing in the \numrange{0.72}{0.8} $F_1$ score range.

The first row of \cref{fig:pr_overall} shows how the performance of the models varies with different choices of the binarization threshold in the grayscale predictions. Observe from \cref{tab:results_synth} that several models exhibit significant discrepancies between their binary $F_1$ scores and the optimal $F_1$ scores obtained from the grayscale predictions. This implies that the models may benefit greatly from employing a dedicated binarization threshold selection algorithm.

It is interesting to see that the models' performance, in general, matched well between the MOBIUS data and MOBIUS-like synthetic data (SynMOBIUS), demonstrating a considerable correspondence between the generated synthetic images and the original real-world data. $F_1$ scores of over $0.8$ for the best performing model indicate that it is possible to learn a competitive segmentation model even from synthetic data with quite different characteristics than the test data. When looking at the performance on SMD+SLD, we observe consistently weaker results for all models compared to the respective performance on MOBIUS or SynMOBIUS, suggesting that the domain shift between the synthetic training data and the SMD+SLD images significantly impacts segmentation results - even for the best performing models. % missmatch was more consistent across the different models, implying that it was a more difficult dataset for the better-performing models to achieve a performance boost on.

%\label{sec:results_mixed}

%\begin{table}%  HMean only
\begin{table*}%  Full
    \caption{\textbf{Comparative assessment of the models trained on mixed data.} The submissions are ranked according to the harmonic means of the achieved $F_1$ scores over the three evaluation datasets. The $F_1$, Precision, Recall and IoU scores were computed from the submitted binary masks. The optimal $F_1$ score on the precision-recall curve ($F_1^{opt}$) and AUC values were calculated from the probabilistic segmentation predictions. The harmonic means are also reported for the rest of the performance measures in the column next to the individual dataset results. Note that the ranking is quite consistent across performance indicators. \vspace{1mm}}
    \label{tab:results_mixed}
    \renewcommand{\arraystretch}{1.1}
    \centering
    \resizebox{0.94\linewidth}{!}{
    \begin{tabular}{rll cccccccc cccc}%  Full
    %\rowcolors{5}{white}{gray!10}%  HMean only
    %\begin{tabular}{rl cccc cc}%  HMean only
        %\hline\hline
        \toprule
        \multirow{2}{*}{Rank} & \multirow{2}{*}{Segmentation Model} & \multirow{2}{*}{Evaluation Dataset} & \multicolumn{8}{c}{From binary masks} & \multicolumn{4}{c}{From probabilistic predictions}\\\cmidrule(lr){4-11} \cmidrule(lr){12-15}%  Full
        & & & \multicolumn{2}{c}{$F_1$} & \multicolumn{2}{c}{Precision} & \multicolumn{2}{c}{Recall} & \multicolumn{2}{c}{IoU} & \multicolumn{2}{c}{$F_1^{opt}$} & \multicolumn{2}{c}{AUC}\\\hline%  Full
        %\multirow{2}{*}{Rank} & \multirow{2}{*}{Segmentation Model} & \multicolumn{4}{c}{Binary} & \multicolumn{2}{c}{Probabilistic}\\\cmidrule(lr){3-6} \cmidrule(lr){7-8}%  HMean only
        %& & $F_1$ & Precision & Recall & IoU & $F_1^{opt}$ & AUC\\\hline%  HMean only
        \multirow{3}{*}{1} & \multirow{3}{*}{SAM-Iris} & MOBIUS & \num{0.850} & \multirow{3}{*}{\num{0.839}} & \num{0.914} & \multirow{3}{*}{\num{0.853}} & \num{0.806} & \multirow{3}{*}{\num{0.840}} & \num{0.751} & \multirow{3}{*}{\num{0.733}} & \num{0.876} & \multirow{3}{*}{\num{0.865}} & \num{0.918} & \multirow{3}{*}{\num{0.910}} \\
 & & SMD+SLD   & \num{0.804} & & \num{0.749} & & \num{0.893} & & \num{0.684} & & \num{0.833} & & \num{0.881} & \\
 & & SynMOBIUS &  \num{0.865} & & \num{0.920} & & \num{0.825} & & \num{0.769} & & \num{0.889} & & \num{0.933} & \\\hline
\multirow{3}{*}{1} & \multirow{3}{*}{ShapeGAN-DLV3+} & MOBIUS & \num{0.858} & \multirow{3}{*}{\num{0.838}} & \num{0.870} & \multirow{3}{*}{\num{0.808}} & \num{0.867} & \multirow{3}{*}{\num{0.889}} & \num{0.769} & \multirow{3}{*}{\num{0.733}} & \num{0.882} & \multirow{3}{*}{\num{0.863}} & \num{0.926} & \multirow{3}{*}{\num{0.909}} \\
 & & SMD+SLD   & \num{0.780} & & \num{0.712} & & \num{0.885} & & \num{0.649} & & \num{0.811} & & \num{0.862} & \\
 & & SynMOBIUS &  \num{0.884} & & \num{0.862} & & \num{0.917} & & \num{0.800} & & \num{0.901} & & \num{0.944} & \\\hline
\multirow{3}{*}{3} & \multirow{3}{*}{SAM2-UNet} & MOBIUS & \num{0.847} & \multirow{3}{*}{\num{0.826}} & \num{0.860} & \multirow{3}{*}{\num{0.795}} & \num{0.858} & \multirow{3}{*}{\num{0.881}} & \num{0.756} & \multirow{3}{*}{\num{0.720}} & \num{0.871} & \multirow{3}{*}{\num{0.852}} & \num{0.906} & \multirow{3}{*}{\num{0.887}} \\
 & & SMD+SLD   & \num{0.775} & & \num{0.704} & & \num{0.888} & & \num{0.649} & & \num{0.806} & & \num{0.842} & \\
 & & SynMOBIUS &  \num{0.862} & & \num{0.841} & & \num{0.899} & & \num{0.767} & & \num{0.882} & & \num{0.918} & \\\hline
\multirow{3}{*}{4} & \multirow{3}{*}{SwinDANet} & MOBIUS & \num{0.848} & \multirow{3}{*}{\num{0.822}} & \num{0.873} & \multirow{3}{*}{\num{0.800}} & \num{0.842} & \multirow{3}{*}{\num{0.864}} & \num{0.756} & \multirow{3}{*}{\num{0.715}} & \num{0.875} & \multirow{3}{*}{\num{0.850}} & \num{0.916} & \multirow{3}{*}{\num{0.896}} \\
 & & SMD+SLD   & \num{0.752} & & \num{0.681} & & \num{0.866} & & \num{0.623} & & \num{0.786} & & \num{0.833} & \\
 & & SynMOBIUS &  \num{0.878} & & \num{0.881} & & \num{0.886} & & \num{0.790} & & \num{0.896} & & \num{0.945} & \\\hline
\multirow{3}{*}{5} & \multirow{3}{*}{UL-VMUNet} & MOBIUS & \num{0.835} & \multirow{3}{*}{\num{0.808}} & \num{0.902} & \multirow{3}{*}{\num{0.804}} & \num{0.796} & \multirow{3}{*}{\num{0.830}} & \num{0.732} & \multirow{3}{*}{\num{0.690}} & \num{0.873} & \multirow{3}{*}{\num{0.838}} & \num{0.927} & \multirow{3}{*}{\num{0.893}} \\
 & & SMD+SLD   & \num{0.739} & & \num{0.673} & & \num{0.845} & & \num{0.600} & & \num{0.770} & & \num{0.820} & \\
 & & SynMOBIUS &  \num{0.859} & & \num{0.881} & & \num{0.850} & & \num{0.760} & & \num{0.881} & & \num{0.942} & \\\hline
\multirow{3}{*}{5} & \multirow{3}{*}{SEG-U-Sclera} & MOBIUS & \num{0.823} & \multirow{3}{*}{\num{0.807}} & \num{0.855} & \multirow{3}{*}{\num{0.799}} & \num{0.815} & \multirow{3}{*}{\num{0.841}} & \num{0.722} & \multirow{3}{*}{\num{0.699}} & \num{0.843} & \multirow{3}{*}{\num{0.833}} & \num{0.834} & \multirow{3}{*}{\num{0.820}} \\
 & & SMD+SLD   & \num{0.778} & & \num{0.729} & & \num{0.862} & & \num{0.659} & & \num{0.811} & & \num{0.786} & \\
 & & SynMOBIUS &  \num{0.823} & & \num{0.823} & & \num{0.846} & & \num{0.721} & & \num{0.846} & & \num{0.840} & \\\hline
\multirow{3}{*}{7} & \multirow{3}{*}{UNet++\_Binary} & MOBIUS & \num{0.811} & \multirow{3}{*}{\num{0.803}} & \num{0.906} & \multirow{3}{*}{\num{0.849}} & \num{0.764} & \multirow{3}{*}{\num{0.792}} & \num{0.710} & \multirow{3}{*}{\num{0.692}} & \num{0.889} & \multirow{3}{*}{\num{0.872}} & \num{0.941} & \multirow{3}{*}{\num{0.928}} \\
 & & SMD+SLD   & \num{0.769} & & \num{0.764} & & \num{0.812} & & \num{0.641} & & \num{0.835} & & \num{0.892} & \\
 & & SynMOBIUS &  \num{0.832} & & \num{0.893} & & \num{0.801} & & \num{0.730} & & \num{0.895} & & \num{0.954} & \\\hline
\multirow{3}{*}{8} & \multirow{3}{*}{AEOS} & MOBIUS & \num{0.850} & \multirow{3}{*}{\num{0.797}} & \num{0.875} & \multirow{3}{*}{\num{0.816}} & \num{0.842} & \multirow{3}{*}{\num{0.799}} & \num{0.754} & \multirow{3}{*}{\num{0.682}} & \num{0.874} & \multirow{3}{*}{\num{0.823}} & \num{0.934} & \multirow{3}{*}{\num{0.880}} \\
 & & SMD+SLD   & \num{0.701} & & \num{0.714} & & \num{0.720} & & \num{0.569} & & \num{0.733} & & \num{0.782} & \\
 & & SynMOBIUS &  \num{0.859} & & \num{0.881} & & \num{0.850} & & \num{0.762} & & \num{0.880} & & \num{0.945} & \\\hline
\multirow{3}{*}{9} & \multirow{3}{*}{KU-CVML} & MOBIUS & \num{0.795} & \multirow{3}{*}{\num{0.780}} & \num{0.725} & \multirow{3}{*}{\num{0.688}} & \num{0.928} & \multirow{3}{*}{\num{0.937}} & \num{0.689} & \multirow{3}{*}{\num{0.658}} & - & \multirow{3}{*}{-} & - & \multirow{3}{*}{-} \\
 & & SMD+SLD   & \num{0.719} & & \num{0.601} & & \num{0.928} & & \num{0.576} & & - & & - & \\
 & & SynMOBIUS &  \num{0.835} & & \num{0.758} & & \num{0.954} & & \num{0.730} & & - & & - &
 \\%  Full
        %\input{Mixed - HMean} \\%  HMean only
        %\hline\hline
        \bottomrule
        \multicolumn{15}{l}{The probabilistic results for KU-CVML are not reported due to issues in the submission.}%  Full
        %\hiderowcolors%  HMean only
        %\multicolumn{8}{l}{The probabilistic results for KU-CVML are not reported due to issues in the submission.}%  HMean only
    \end{tabular}%
    }
%\end{table}%  HMean only
\vspace{-3mm}
\end{table*}%  Full

\vspace{1.2mm}\noindent\textbf{Mixed Track.}~In the mixed track, participants could use both the synthetic as well as real-world data for model training. As can be seen from \cref{fig:bar_mixed,tab:results_mixed}, the results of this track were closer (considering the strongest and weakest model) than those of the synthetic track. This suggests that it is still easier to train a well performing model with a mix of real and synthetic data than with synthetic data alone, where dedicated mechanisms need to be incorporated into the learning procedure to account for the domain shift w.r.t. real-world data. The winner of the mixed track, \textbf{SAM-Iris}, and the runner-up, \textbf{ShapeGAN-DLV3+}, were within $0.001$ of each other in terms of their $F_1$ scores, which is well within the margin of error. As such, they are considered the joint winners of the SSBC 2025 Mixed track.~Notably, all the approaches resulted in competitive performance, with $F_1$ scores in the range of \numrange{0.78}{0.84}.

It is interesting to note that the two worst performers from the Synthetic track performed the best in the Mixed track, driven chiefly by their significant performance improvement on the SMD+SLD evaluation data. The two best performers from the Synthetic track maintained their high performance, taking the \nth{3} and \nth{4} place in the Mixed track. %KU-CVML (3rd in the Synthetic track), on the other hand, dropped to the last place in the Mixed track, largely due to its poor performance on SMD+SLD data, which, unlike the other models, it did not manage to improve relative to the Synthetic track.

The second row of \cref{fig:pr_overall} shows the Mixed track results over the grayscale predictions. We observe the significantly tighter precision-recall curves relative to the Synthetic track in the top row, again implying a much closer performance of the submitted models. The discrepancies between the binary $F_1$ and optimal grayscale $F_1$ scores are still present, but significantly lower. This implies that a good thresholding strategy is of less importance with mixed-trained models. The exception is UNet++\_Binary, which achieves the top performance in its probabilistic metrics, but only the \nth{7} place in the binary ranking.

\subsection{Differential Performance}
\label{sec:results_delta}

The goal of SSBC 2025 was to establish how viable the use of synthetic data is in developing sclera segmentation models. As such, we are interested in the performance differentials between the models trained on the synthetic data and their mixed-trained counterparts.

\begin{figure}[t]
    \centering
    \includegraphics[width=\linewidth]{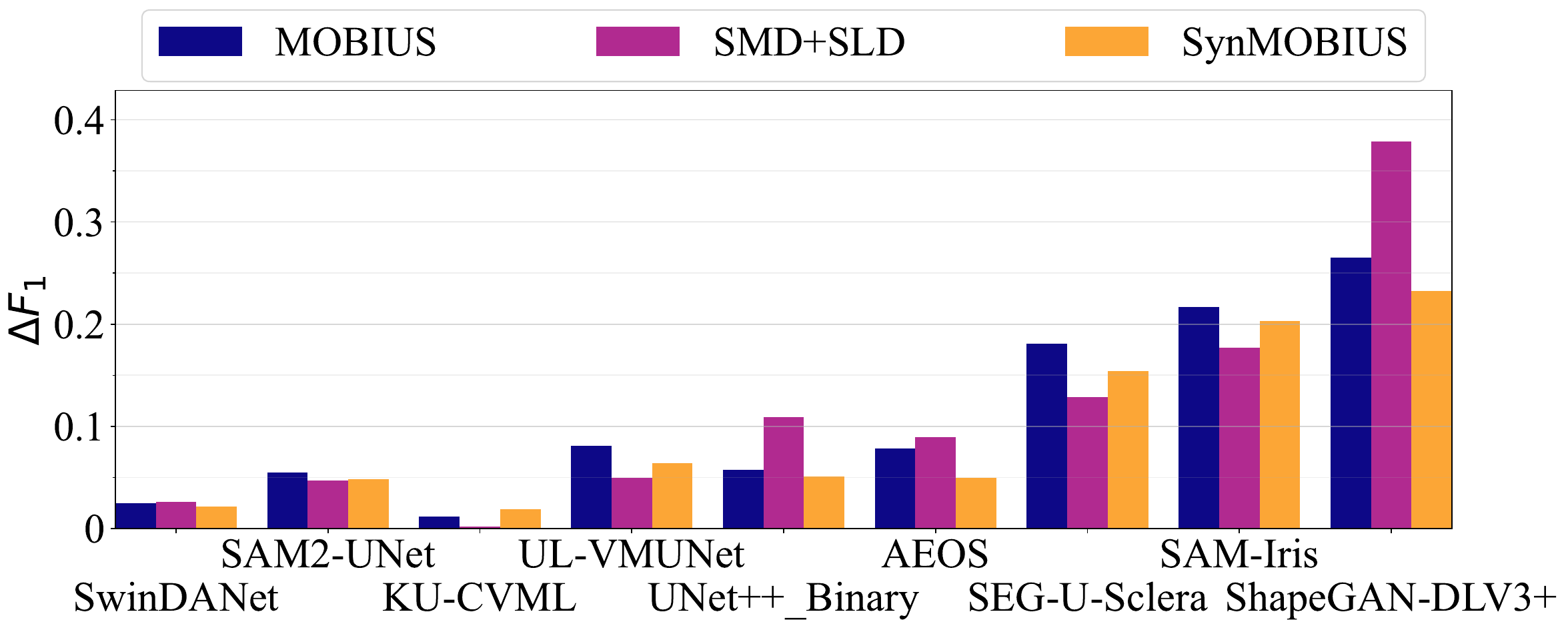}
    \caption{\textbf{Performance differentials} of the submitted models from the Synthetic track to the Mixed track on three test datasets, i.e., MOBIUS, SMD+SLD and SynMOBIUS. The graph shows differences in the achieved $F_1$ scores, i.e., $F_1^{mixed}-F_1^{synthetic}$. Positive values denote a better performance in the Mixed track.\vspace{-2mm}}
    \label{fig:bar_delta}
\end{figure}

From \cref{fig:bar_delta}, we can see that every single model's performance improved from the Synthetic to the Mixed track. However, for most of the models, the performance boost was on the lower end, mostly under $0.05$ in terms of $F_1$ score. However, for the three worst-performing models of the Synthetic track, the performance boost from adding the real-world training data was substantial. This implies that the choice of architecture is important when training on synthetic data alone, as certain model architectures perform significantly worse without real-world samples to complement the synthetic data. However, with many of the performance boosts being small, the synthetic training data has shown to be a feasible approach for training most models in cases where privacy protection is essential.

\begin{figure}[t]
    \centering
    \begin{subfigure}{.36\linewidth}
        \includegraphics[width=\textwidth]{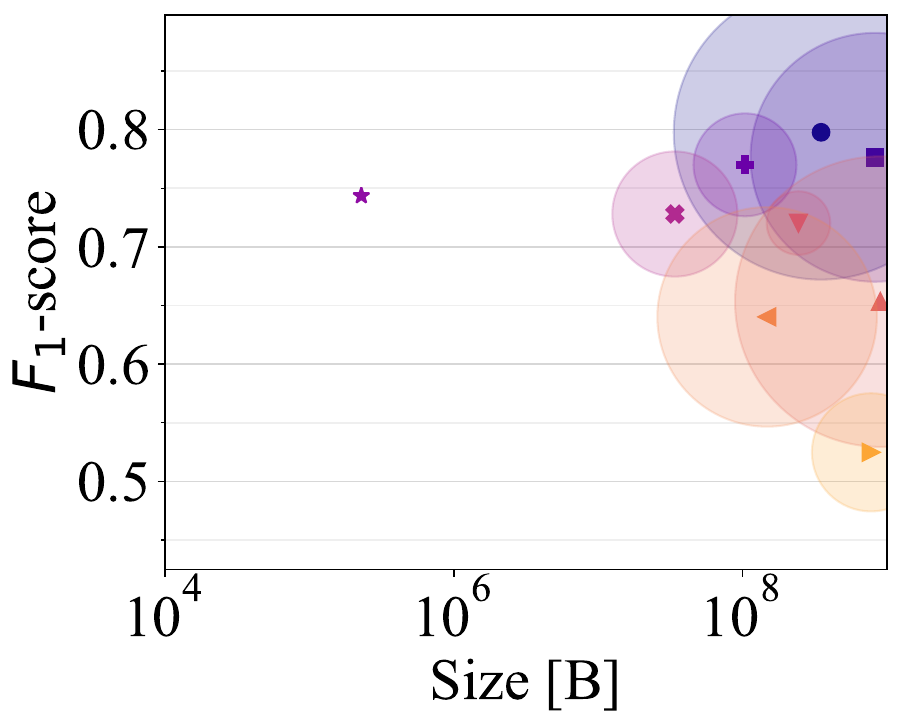}
        \caption{Synthetic}
        \label{fig:sc_synth}
    \end{subfigure}\hfill%
    \begin{subfigure}{.62\linewidth}
        \includegraphics[width=\textwidth]{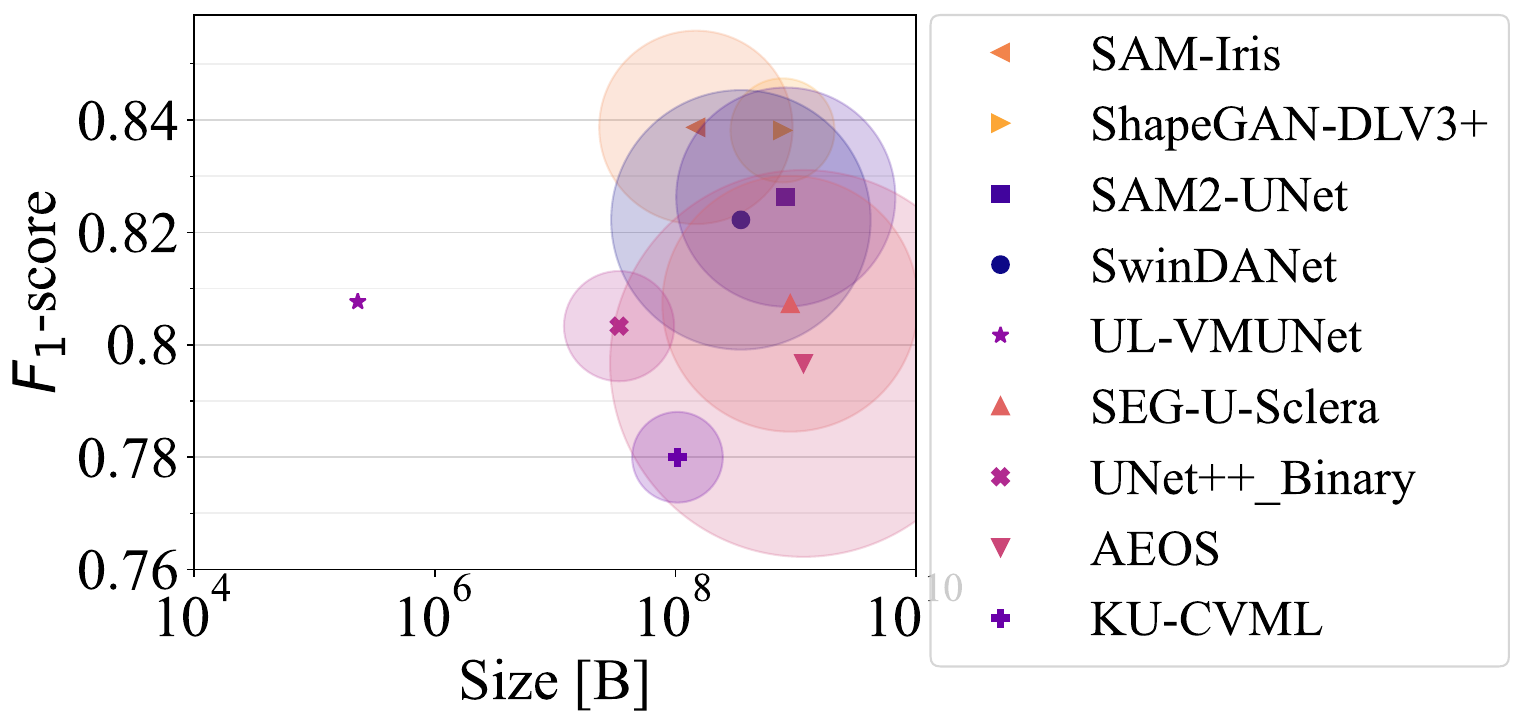}
        \caption{Mixed}
        \label{fig:sc_mixed}
    \end{subfigure}
    \caption{\textbf{Performance/complexity trade-off}. The areas of the circles correspond to the computational complexity of the models, expressed in FLOPs. Best viewed in color and zoomed in. \vspace{-3mm}}
    \label{fig:sc_overall}
\end{figure}

\begin{figure*}[t]
    \centering
    \begin{subfigure}{.09\linewidth}
        \includegraphics[width=\textwidth,height=.57\textwidth]{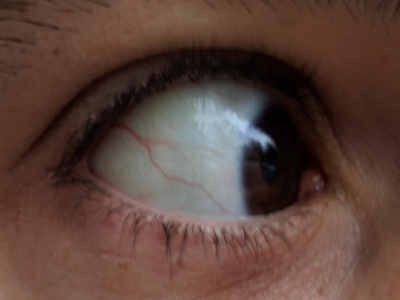}\\
        \includegraphics[width=\textwidth,height=.57\textwidth]{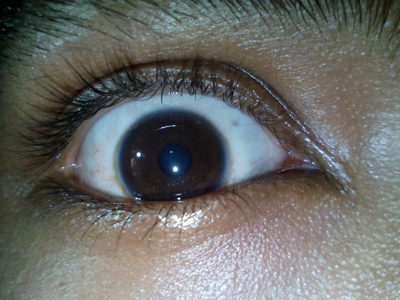}\\
        \includegraphics[width=\textwidth,height=.57\textwidth]{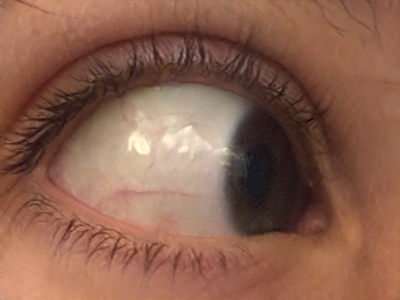}\\[1mm]
        \includegraphics[width=\textwidth,height=.57\textwidth]{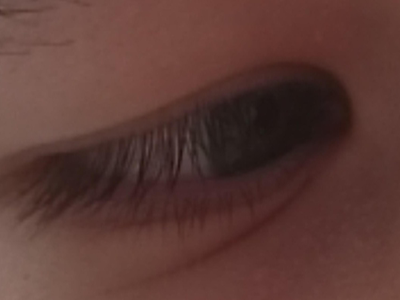}\\
        \includegraphics[width=\textwidth,height=.57\textwidth]{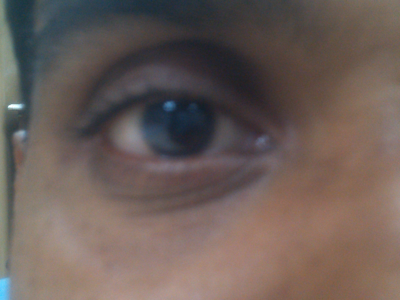}\\
        \includegraphics[width=\textwidth,height=.57\textwidth]{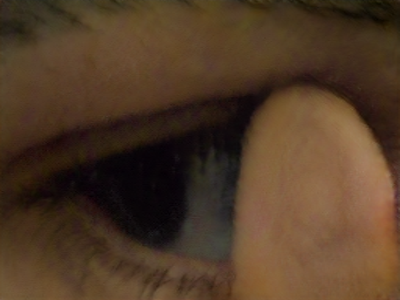}
        \caption{}
        \label{fig:q_img}
    \end{subfigure}\hfill%
    \begin{subfigure}{.09\linewidth}
        \includegraphics[width=\textwidth,height=.57\textwidth]{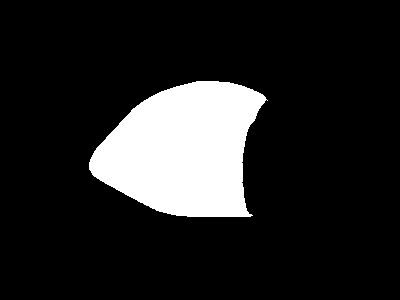}\\
        \includegraphics[width=\textwidth,height=.57\textwidth]{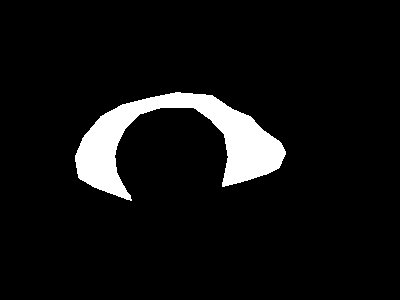}\\
        \includegraphics[width=\textwidth,height=.57\textwidth]{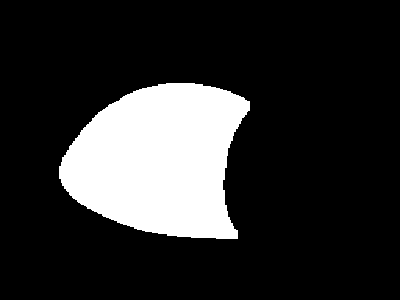}\\[1mm]
        \includegraphics[width=\textwidth,height=.57\textwidth]{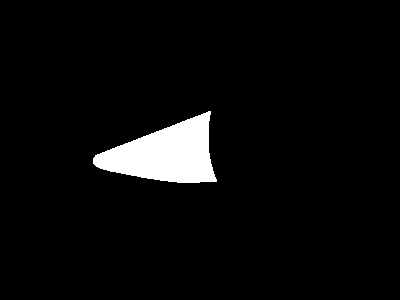}\\
        \includegraphics[width=\textwidth,height=.57\textwidth]{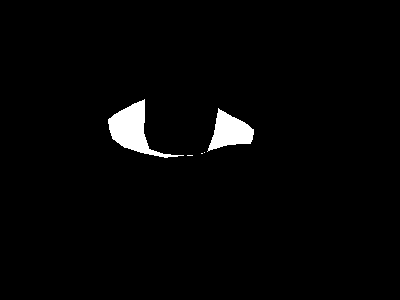}\\
        \includegraphics[width=\textwidth,height=.57\textwidth]{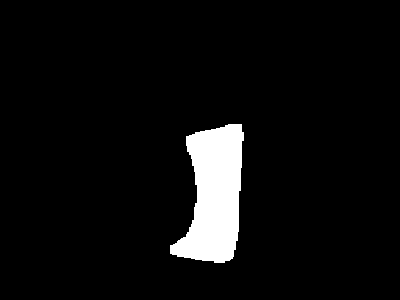}
        \caption{}
        \label{fig:q_gt}
    \end{subfigure}\hfill%
    \begin{subfigure}{.09\linewidth}
        \includegraphics[width=\textwidth,height=.57\textwidth]{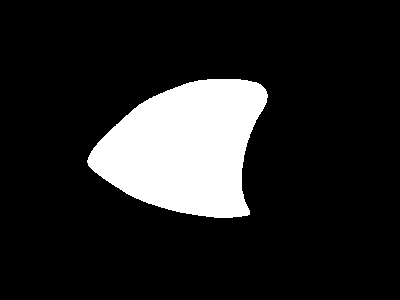}\\
        \includegraphics[width=\textwidth,height=.57\textwidth]{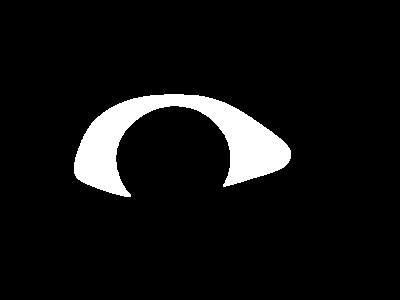}\\
        \includegraphics[width=\textwidth,height=.57\textwidth]{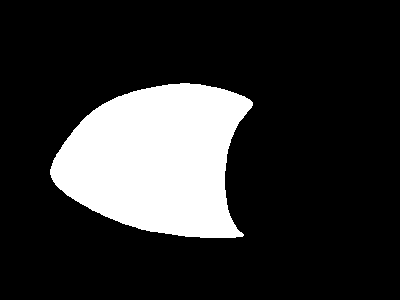}\\[1mm]
        \includegraphics[width=\textwidth,height=.57\textwidth]{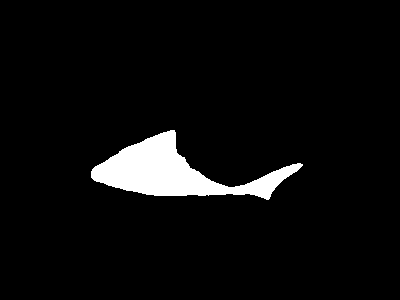}\\
        \includegraphics[width=\textwidth,height=.57\textwidth]{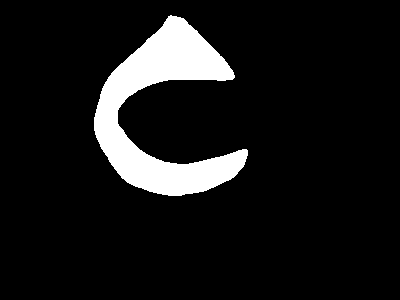}\\
        \includegraphics[width=\textwidth,height=.57\textwidth]{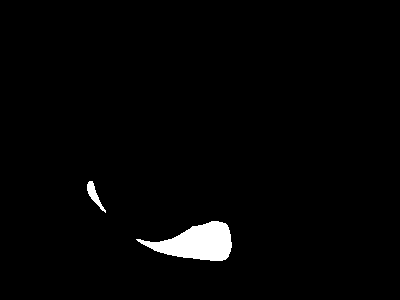}
        \caption{}
        \label{fig:q_swindanet}
    \end{subfigure}\hfill%
    \begin{subfigure}{.09\linewidth}
        \includegraphics[width=\textwidth,height=.57\textwidth]{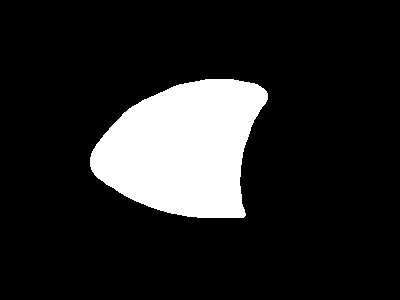}\\
        \includegraphics[width=\textwidth,height=.57\textwidth]{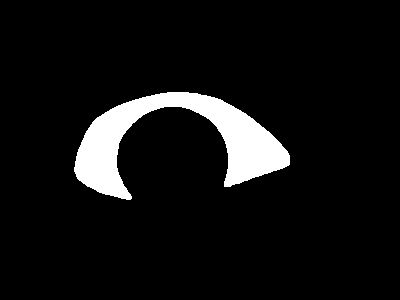}\\
        \includegraphics[width=\textwidth,height=.57\textwidth]{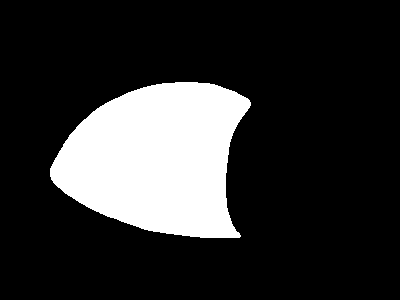}\\[1mm]
        \includegraphics[width=\textwidth,height=.57\textwidth]{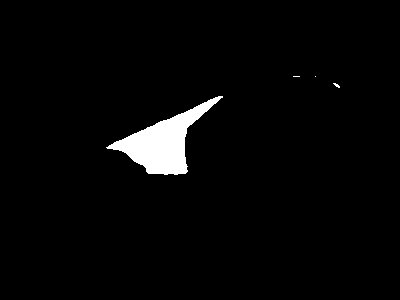}\\
        \includegraphics[width=\textwidth,height=.57\textwidth]{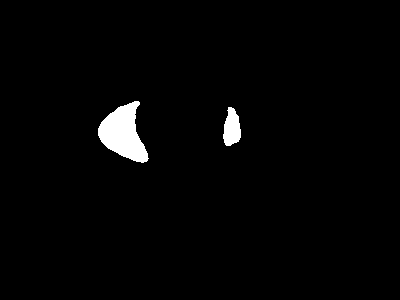}\\
        \includegraphics[width=\textwidth,height=.57\textwidth]{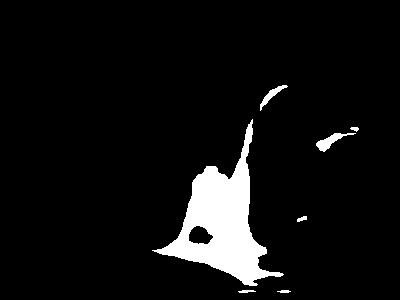}
        \caption{}
        \label{fig:q_sam2unet}
    \end{subfigure}\hfill%
    \begin{subfigure}{.09\linewidth}
        \includegraphics[width=\textwidth,height=.57\textwidth]{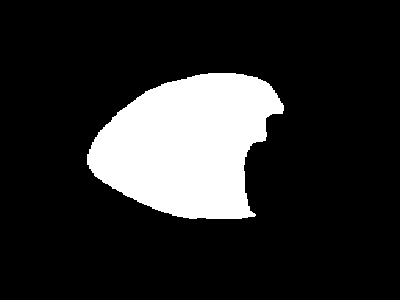}\\
        \includegraphics[width=\textwidth,height=.57\textwidth]{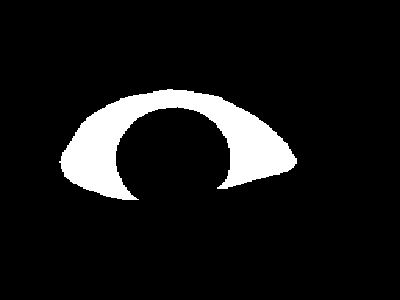}\\
        \includegraphics[width=\textwidth,height=.57\textwidth]{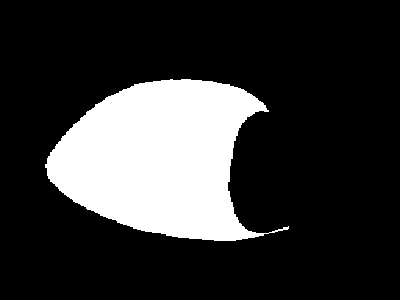}\\[1mm]
        \includegraphics[width=\textwidth,height=.57\textwidth]{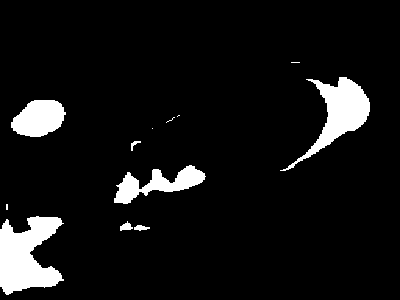}\\
        \includegraphics[width=\textwidth,height=.57\textwidth]{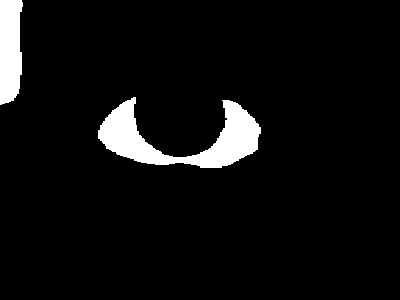}\\
        \includegraphics[width=\textwidth,height=.57\textwidth]{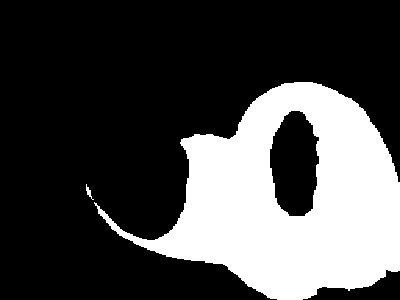}
        \caption{}
        \label{fig:q_kucvml}
    \end{subfigure}\hfill%
    \begin{subfigure}{.09\linewidth}
        \includegraphics[width=\textwidth,height=.57\textwidth]{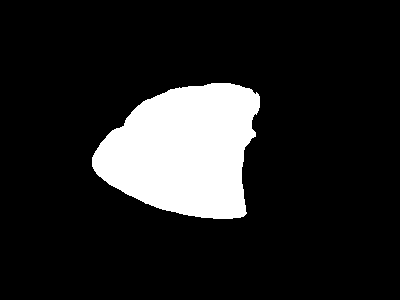}\\
        \includegraphics[width=\textwidth,height=.57\textwidth]{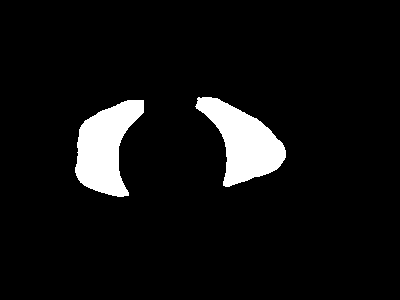}\\
        \includegraphics[width=\textwidth,height=.57\textwidth]{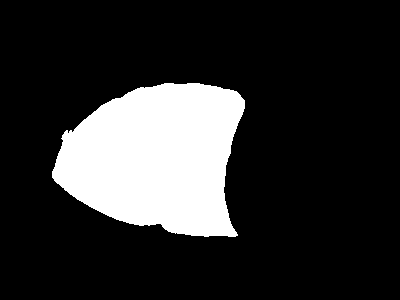}\\[1mm]
        \includegraphics[width=\textwidth,height=.57\textwidth]{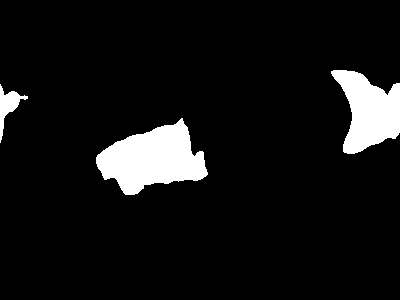}\\
        \includegraphics[width=\textwidth,height=.57\textwidth]{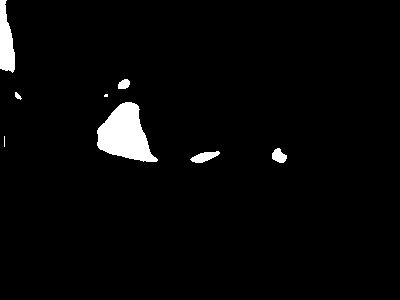}\\
        \includegraphics[width=\textwidth,height=.57\textwidth]{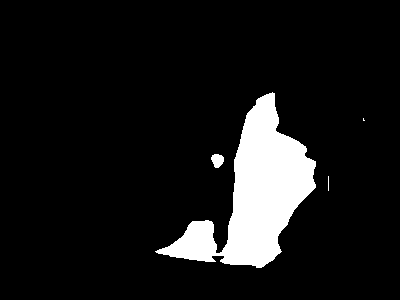}
        \caption{}
        \label{fig:q_ulvmunet}
    \end{subfigure}\hfill%
    \begin{subfigure}{.09\linewidth}
        \includegraphics[width=\textwidth,height=.57\textwidth]{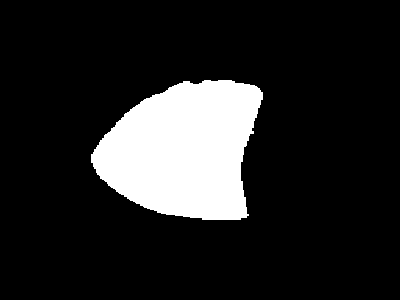}\\
        \includegraphics[width=\textwidth,height=.57\textwidth]{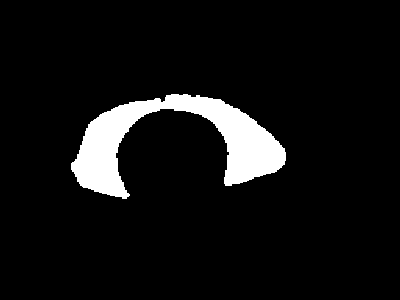}\\
        \includegraphics[width=\textwidth,height=.57\textwidth]{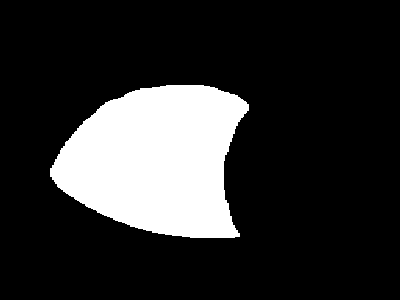}\\[1mm]
        \includegraphics[width=\textwidth,height=.57\textwidth]{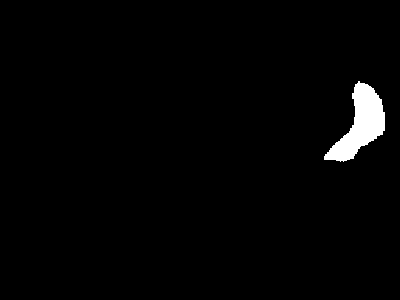}\\
        \includegraphics[width=\textwidth,height=.57\textwidth]{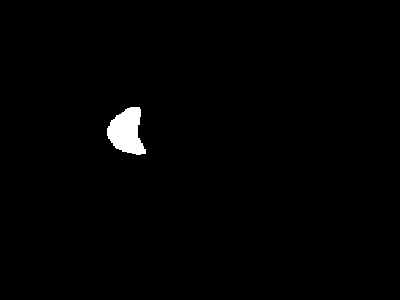}\\
        \includegraphics[width=\textwidth,height=.57\textwidth]{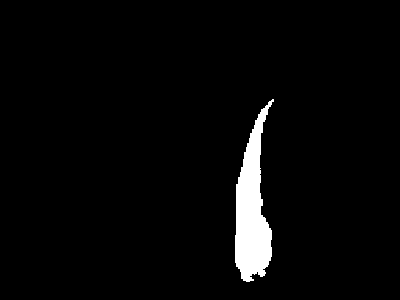}
        \caption{}
        \label{fig:q_unetbinary}
    \end{subfigure}\hfill%
    \begin{subfigure}{.09\linewidth}
        \includegraphics[width=\textwidth,height=.57\textwidth]{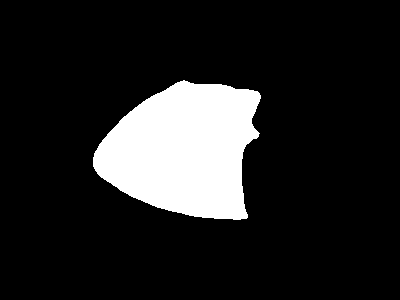}\\
        \includegraphics[width=\textwidth,height=.57\textwidth]{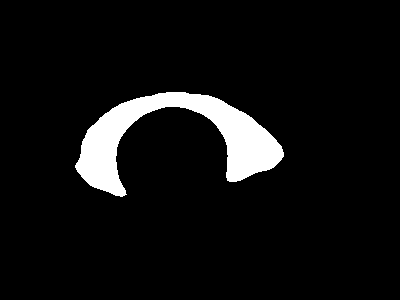}\\
        \includegraphics[width=\textwidth,height=.57\textwidth]{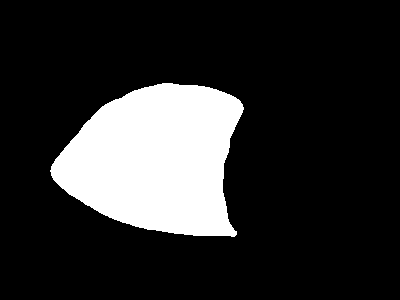}\\[1mm]
        \includegraphics[width=\textwidth,height=.57\textwidth]{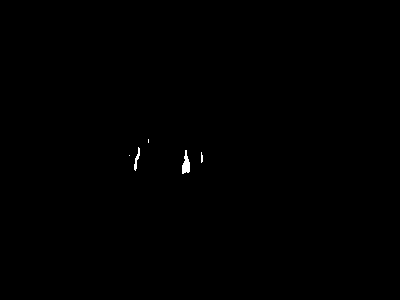}\\
        \includegraphics[width=\textwidth,height=.57\textwidth]{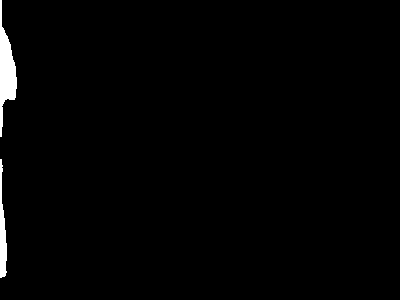}\\
        \includegraphics[width=\textwidth,height=.57\textwidth]{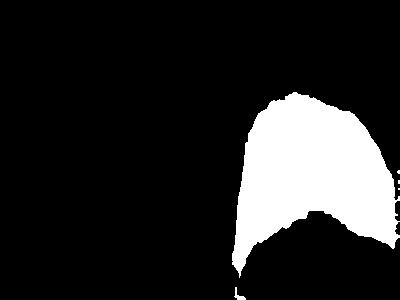}
        \caption{}
        \label{fig:q_aeos}
    \end{subfigure}\hfill%
    \begin{subfigure}{.09\linewidth}
        \includegraphics[width=\textwidth,height=.57\textwidth]{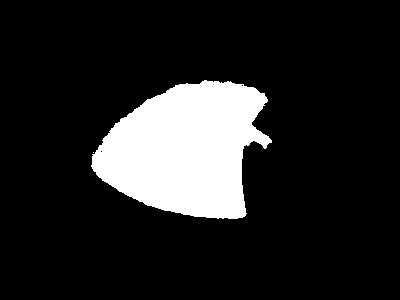}\\
        \includegraphics[width=\textwidth,height=.57\textwidth]{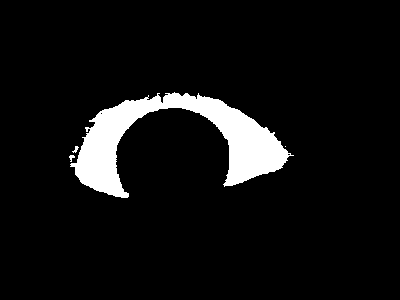}\\
        \includegraphics[width=\textwidth,height=.57\textwidth]{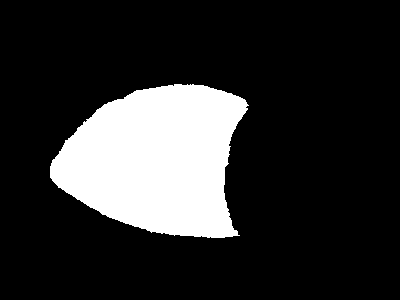}\\[1mm]
        \includegraphics[width=\textwidth,height=.57\textwidth]{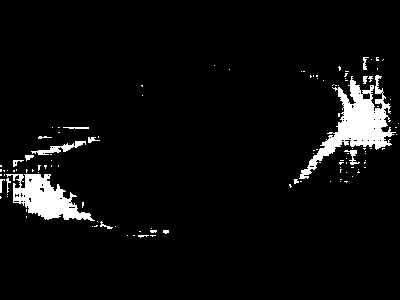}\\
        \includegraphics[width=\textwidth,height=.57\textwidth]{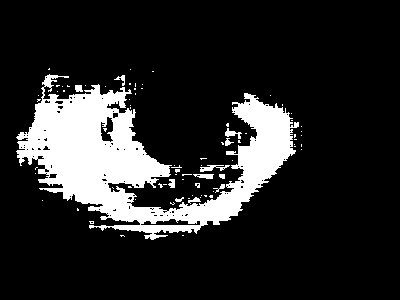}\\
        \includegraphics[width=\textwidth,height=.57\textwidth]{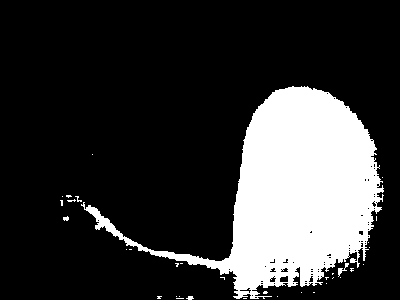}
        \caption{}
        \label{fig:q_segusclera}
    \end{subfigure}\hfill%
    \begin{subfigure}{.09\linewidth}
        \includegraphics[width=\textwidth,height=.57\textwidth]{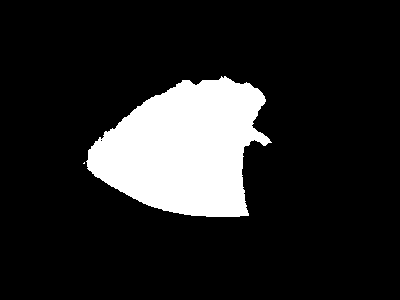}\\
        \includegraphics[width=\textwidth,height=.57\textwidth]{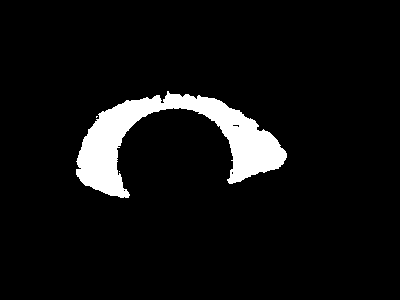}\\
        \includegraphics[width=\textwidth,height=.57\textwidth]{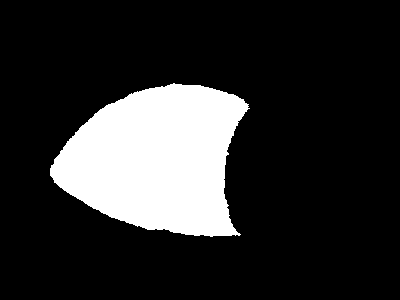}\\[1mm]
        \includegraphics[width=\textwidth,height=.57\textwidth]{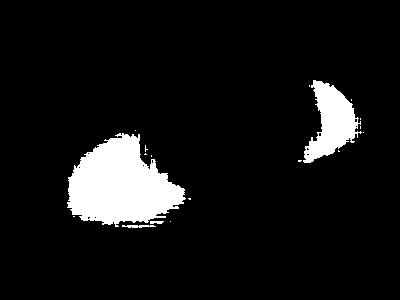}\\
        \includegraphics[width=\textwidth,height=.57\textwidth]{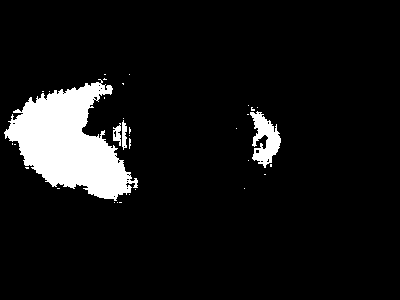}\\
        \includegraphics[width=\textwidth,height=.57\textwidth]{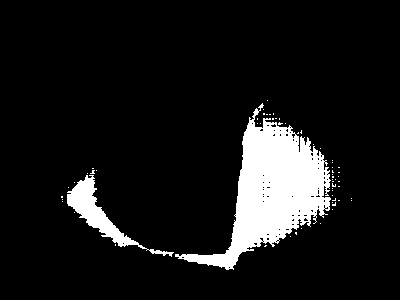}
        \caption{}
        \label{fig:q_samiris}
    \end{subfigure}\hfill%
    \begin{subfigure}{.09\linewidth}
        \includegraphics[width=\textwidth,height=.57\textwidth]{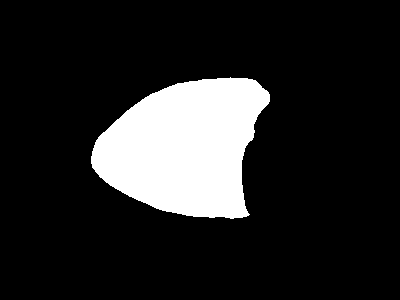}\\
        \includegraphics[width=\textwidth,height=.57\textwidth]{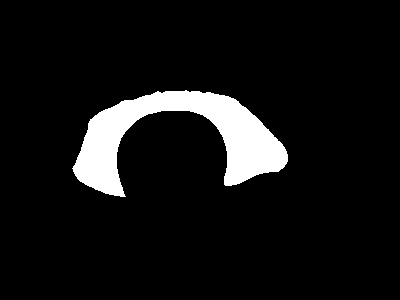}\\
        \includegraphics[width=\textwidth,height=.57\textwidth]{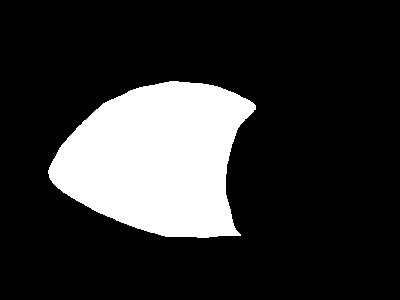}\\[1mm]
        \includegraphics[width=\textwidth,height=.57\textwidth]{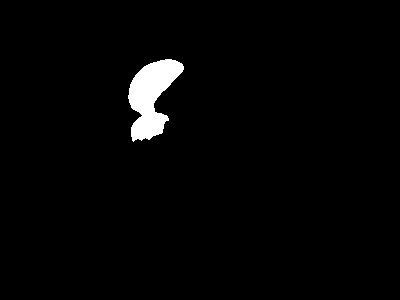}\\
        \includegraphics[width=\textwidth,height=.57\textwidth]{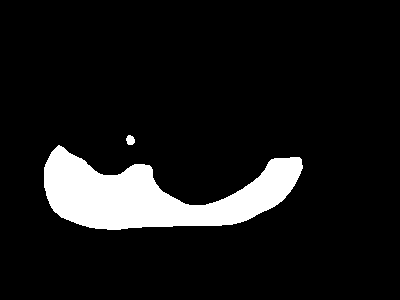}\\
        \includegraphics[width=\textwidth,height=.57\textwidth]{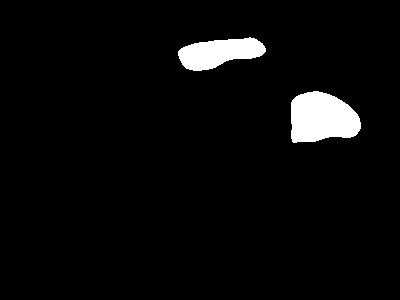}
        \caption{}
        \label{fig:q_shapegandlv3}
    \end{subfigure}\vspace{-1mm}
    \caption{Qualitative comparison of the submitted models (in terms of binary masks) on selected images. The top three rows contain the well-performing samples from the evaluation datasets (in order, MOBIUS, SMD+SLD, SynMOBIUS), while the bottom three rows show poor-performing samples. Observe the difference in the segmentation quality across the evaluated models. The figure shows \subref{fig:q_img} the original image; \subref{fig:q_gt} the ground truth mask; and the submitted binary masks from: \subref{fig:q_swindanet} SwinDANet, \subref{fig:q_sam2unet} SAM2-UNet, \subref{fig:q_kucvml} KU-CVML, \subref{fig:q_ulvmunet} UL-VMUNet, \subref{fig:q_unetbinary} UNet++\_Binary, \subref{fig:q_aeos} AEOS, \subref{fig:q_segusclera} SEG-U-Sclera, \subref{fig:q_samiris} SAM-Iris, and \subref{fig:q_shapegandlv3} ShapeGAN-DLV3+.\vspace{-4mm}}
    \label{fig:qualitative}
\end{figure*}

The largest difference in performance was seen with the ShapeGAN-DLV3+ model, which features a generator-discriminator architecture, albeit different from classical GANs, since the discriminator is attached to the generator bottleneck. Since GANs are well known to require large amounts of training data, we can partly attribute the difference in performance simply to the increased size of the training dataset with additional real-world samples. However, we note that the real-world SBVPI dataset was significantly smaller than the synthetic training dataset.~The second-largest difference came with the SAM-Iris model, which relies on a training pipeline with an explicit transition from synthetic to real training data, which explains the discrepancy in performance.~Similarly, AEOS (which resulted in the fourth-largest difference) employed completely different architectures for the two tracks.~As such, we note that three of the four biggest performance differentials between the Synthetic and the Mixed track are a result of explicit methodological choices rather than domain shifts.

\subsection{Performance Across Different Complexities}
\label{sec:results_complexity}

Finally, we study the trade-off between model performance and complexity, which is a key factor for real-world deployment. From \cref{fig:sc_overall}, we see that most of the submitted models' complexities are roughly on par. However, a notable exception here is UL-VMUnet, which achieved competitive performance despite a comparatively tiny size (229 KB) and computational complexity (60 MFLOPs). This result is in line with recent works in sclera biometrics \cite{vitek2023ipad,vitek2025gazenet}, where it was shown that typically employed network architectures could be significantly reduced in size and complexity without a noticeable degradation in performance.

It is also worth noting that UL-VMUNet ranked higher in the Synthetic than in the Mixed track, which may imply that smaller, less complex models generalize better when trained with synthetic data and are not disturbed as easily by artifacts produced by the data generation procedure.

\subsection{Qualitative Evaluation}
\label{sec:qualitative}

In \cref{fig:qualitative}, we show some of the best and worst cases of the generated segmentation masks. Note that even on the best-performing samples, models often struggled with minor artifacts, such as those resulting from specular reflection in the iris.~Among the poor performing samples, we often find images where the sclera is occluded, either by the eyelid (in the case of a partially closed eye, such as in row 4 of \cref{fig:qualitative}) or by external sources (such as the finger in row 6 of \cref{fig:qualitative}). Another source of errors is lighting-induced skin discolouration (row 5 of \cref{fig:qualitative}), particularly when the skin around the eye appears significantly lighter than elsewhere. Additionally, a poorly-lit sclera (row 6 of \cref{fig:qualitative}) can also cause issues in segmentation with many of the models.

%%%%%%%%%%%%%%%%%%%%%%%%%%%%%%%%%%%%%%%%%%%%%%%%%%%%%%%%%%%%%%%%%%%%%%%%%%%%%%%%%%%%%
\section{Conclusion}
\label{sec:conclusion}
%%%%%%%%%%%%%%%%%%%%%%%%%%%%%%%%%%%%%%%%%%%%%%%%%%%%%%%%%%%%%%%%%%%%%%%%%%%%%%%%%%%%%
The 2025 edition of the Sclera Segmentation Benchmarking Competition (SSBC 2025) was organized to benchmark the performance of contemporary segmentation models in the task of sclera segmentation, and to establish the viability of using synthetically generated data to develop and train such models. The use of synthetic data ensures no identifying information is required to train segmentation models and facilitates the ethical development of biometric systems. To this end, SSBC 2025 was conducted in two tracks, which differed in the training data the contestants used -- purely synthetic data for the first track and a mix of synthetic and real-world data for the second. Nine research groups participated in the competition.

The winner of the Synthetic track was \textbf{SwinDANet} (submitted by the Couger Inc.\ team), while the Mixed track had two joint winners, whose results were within 0.001 of each other: \textbf{SAM-Iris} (Idiap -- HES-SO team) and \textbf{ShapeGAN-DLV3+} (AU team). The results of the competition point to the high fidelity of synthetic data and its viability in model training, as most approaches performed roughly as well in the Synthetic track as in the Mixed one, and the models that did substantially improve their performance between the tracks, did so mostly due to methodological decisions, rather than any infidelities or improprieties in the synthetic training data. This is an important result for future research, as synthetically generated datasets enable biometric research without the risks of privacy breaches or abuse, as well as being significantly easier to obtain or compile.
%%%%%%%%%%%%%%%%%%%%%%%%%%%%%%%%%%%%%%%%%%%%%%%%%%%%%%%%%%%%%%%%%%%%%%%%%%%%%%%%%%%%%

%%%%%%%%%%%%%%%%%%%%%%%%%%%%%%%%%%%%%%%%%%%%%%%%%%%%%%%%%%%%%%%%%%%%%%%%%%%%%%%%%%%%%
%\color{blue}
%\section*{}

{%\small%
%\vspace{1mm}\noindent\textbf{Acknowledgments.}~
\section*{Acknowledgments}
Supported in parts by the ARIS Research Programmes P2-0250 Metrology and Biometric Systems, P2-0214 Computer Vision, the ARIS young researcher program, and the SNSF Projects 214653 FairMI and 213369 StreamKG. %by the European Union’s Horizon 2020  research and innovation programme under the Marie Skłodowska-Curie grant agreement No. 860813 - TReSPAsS-ETN and the German Federal Ministry of Education and Research and the Hessen State Ministry for Higher Education, Research and the Arts within their joint support of the National Research Center for Applied Cybersecurity ATHENE.
}

%%%%%%%%%%%%%%%%%%%%%%%%%%%%%%%%%%%%%%%%%%%%%%%%%%%%%%%%%%%%%%%%%%%%%%%%%%%%%%%%%%%%%

{\small
\bibliographystyle{ieee}
\bibliography{Literature}
}

\end{document}